\pdfoutput=1
\documentclass[11pt]{article}

\usepackage{EMNLP2022}

\usepackage{times}
\usepackage{latexsym}

\usepackage[T1]{fontenc}
\usepackage[utf8]{inputenc}

\usepackage{microtype}

\usepackage{inconsolata}

\usepackage{times}
\usepackage{latexsym}

\usepackage[T1]{fontenc}
\usepackage[utf8]{inputenc}

\usepackage{microtype}

\usepackage{graphicx}
\usepackage{amsmath}
\usepackage{booktabs}
\usepackage{nicefrac}
\usepackage{enumitem}
\usepackage{sidecap}
\usepackage{multirow}
\usepackage{marvosym}
\usepackage{amsmath,graphicx}
\usepackage{amssymb}
\usepackage{amsthm}
\usepackage{mathtools}
\usepackage{bm}
\usepackage{bbm}

\title{Chunk-based Nearest Neighbor Machine Translation}

\author{Pedro Henrique Martins\textsuperscript{\Neptune\Scorpio} \quad
        Zita Marinho\textsuperscript{\Saturn\Moon} \quad
        Andr\'e F.~T. Martins\textsuperscript{\Neptune\Pluto\Scorpio} \\
\textsuperscript{\Neptune}Instituto de Telecomunica\c{c}\~oes~ \quad
\textsuperscript{\Scorpio}Unbabel~ \quad
\textsuperscript{\Saturn}Institute of Systems and Robotics~\\
\textsuperscript{\Moon}DeepMind~ \quad \textsuperscript{\Pluto}LUMLIS (Lisbon ELLIS Unit), Instituto Superior T\'ecnico~ \\
Lisbon, Portugal\\
\href{mailto:pedrohenriqueamartins@tecnico.ulisboa.pt}{\tt pedrohenriqueamartins@tecnico.ulisboa.pt},\\
\href{mailto:zmarinho@google.com}{\tt zmarinho@google.com}, \quad
\href{andre.t.martins@tecnico.ulisboa.pt}{\tt andre.t.martins@tecnico.ulisboa.pt}.
}

\begin{document}
\maketitle

\begin{abstract}
Semi-parametric models, which augment generation with retrieval, have led to impressive results in language modeling and machine translation, due to their ability to retrieve fine-grained information from a datastore of examples.
One of the most prominent approaches, $k$NN-MT, exhibits strong domain adaptation capabilities by retrieving tokens from domain-specific datastores  \citep{khandelwal2020nearest}. However, $k$NN-MT requires an expensive retrieval operation for every single generated token, leading to a very low decoding speed (around 8 times slower than a parametric model).
In this paper, we introduce a \textit{chunk-based} $k$NN-MT model which 
retrieves chunks of tokens from the datastore, instead of a single token. We propose several strategies 
for incorporating the retrieved chunks into the generation process, and for selecting the steps at which the model needs to search for neighbors in the datastore.
Experiments on machine translation in two settings, static and ``on-the-fly'' domain adaptation, show that the chunk-based $k$NN-MT model leads to significant speed-ups (up to 4 times) with only a small drop in translation quality.\footnote{The code is available at \url{https://github.com/deep-spin/chunk-based_knn-mt}.}
\end{abstract}

\section{Introduction}
Machine translation has seen remarkable advances due to increasingly powerful neural models \citep{sutskever2014sequence, bahdanau2015neural, vaswani2017attention}. 
Most deployed systems are \textit{fully-parametric} (the training data is fully compressed into the parameters of a neural model), but they often struggle when translating rare words or out-of-domain sentences \citep{koehn2017six}, commonly requiring several stages of fine-tuning to adapt to data drift or to new domains. 
Recently, \textbf{semi-parametric methods} have shown great promise, by combining the strengths of parametric models with external databases of parallel sentences, such as translation memories  \citep{gu2018search, zhang2018guiding, bapna2019non, meng2021fast, zheng2021adaptive, jiang2021learning, martins2022efficient}. 

One of the most prominent semi-parametric models for machine translation is the $k$-Nearest Neighbor Machine Translation model ($k$NN-MT) \citep{khandelwal2020nearest}, which has led to impressive results, particularly in domain adaptation settings, without requiring fine-tuning. 
The $k$NN-MT model constructs domain-specific datastores of parallel sentences and, at inference time, it  retrieves similar examples from these datastores, which are used to improve the generation process, through the interpolation of probability distributions. 
However, $k$NN-MT only retrieves single tokens---this is inefficient, since the model needs to consult the datastore at every generation step, an expensive operation. Consequently, its decoding speed is around 8 times slower than that of a fully-parametric model.

Recent work has introduced several techniques to speed up $k$NN-MT. \citet{meng2021fast} proposed  Fast $k$NN-MT, which constructs a different datastore for each example, by first searching for the nearest neighbors of the source tokens. \citet{wang2021faster} introduced Faster $k$NN-MT, similar to Fast $k$NN-MT but with reduced memory requirements. 
\citet{martins2022efficient} proposed pruning the datastore, reducing the keys' representation size, and using a cache of retrieval distributions.
However, despite leading to some decoding speedups, these methods are limited as they still \textbf{retrieve a single token at each time step}.

In this paper, we propose a simple and efficient chunk-based $k$NN-MT model. Inspired by RETRO \citep{borgeaud2021improving}, the chunk-based $k$NN-MT model retrieves \textit{chunks} of tokens, instead of single tokens. But, similarly to $k$NN-MT and unlike RETRO, \textbf{it does not require any training or fine-tuning of the parametric component}: it simply uses a combination of caching and interpolation of probability distributions to incorporate the retrieved tokens.
By doing this, the model leads to a similar translation quality while having to search for neighbors in the datastore less often. This leads to decoding speeds \textbf{up to 4 times faster} than the ones achieved using the vanilla $k$NN-MT model and only twice as slow as a fully-parametric model, but with considerably higher translation quality.

In sum, our main contributions are:
\begin{itemize}
    \item We introduce a chunk-based $k$NN-MT model, which retrieves chunks of tokens from a datastore of examples. 
    \item We propose and compare several approaches to deal with the retrieved chunks' tokens and to select the steps in which the model performs retrieval from the datastore.
    \item We compare the translation quality and decoding efficiency on domain adaptation, which shows the benefits of chunk-based $k$NN-MT.
    \item We propose using chunk-based $k$NN-MT for on-the-fly adaptation. 
\end{itemize}

\section{Background}
In machine translation, a model is given a sentence or document in a source language, $\bm{x}=[x_1, \dots, x_L]$, and the goal is to output a translation in a target language, $\bm{y}=[y_1, \dots, y_N]$. 
This is commonly done using a parametric sequence-to-sequence model \citep{sutskever2014sequence,bahdanau2015neural,vaswani2017attention}, in which the encoder receives the source sentence as input and outputs a set of hidden states. Then, at each step $t$, the decoder attends to these hidden states and outputs a probability distribution over the vocabulary, $p_{\mathrm{NMT}}(y_t|\bm{y}_{<t}, \bm{x})$. 
Finally, these probability distributions are used in a search procedure to generate the translation, typically using beam search \citep{reddy1077}.

\subsection{$k$-Nearest Neighbor Machine Translation}
\citet{khandelwal2020nearest} introduced a semi-parametric model called \textbf{$k$-nearest neighbor machine translation} ($k$NN-MT). $k$NN-MT is composed of a parametric component that outputs a probability distribution over the vocabulary as above, $p_{\mathrm{NMT}}(y_t|\bm{y}_{<t}, \bm{x})$,  enhanced with an approximate nearest neighbor retrieval mechanism, which allows direct access to a datastore of examples.

The $k$NN-MT's datastore $\mathcal{D}$ consists of a key-value memory, where each key is the decoder's output representation, $\bm{f}(\bm{x},\bm{y}_{<t}) \in \mathbb{R}^d$, and the value is the corresponding target token $y_t \in \mathcal{V}$:
\begin{equation}
    \mathcal{D} = \left\{\left(\bm{f}(\bm{x},\bm{y}_{<t}\right),y_t) \; \forall \; t \mid (\bm{x},\bm{y})  \in \mathcal{S}\right\},
\end{equation}
where $\mathcal{S}$ denotes a set of parallel sentences.
Then, at inference time, the model searches the datastore to (approximately) retrieve the set of $k$ nearest neighbors $\mathcal{N}$. 
The retrieval distribution, $p_{k\mathrm{NN}}(y_t|\bm{y}_{<t}, \bm{x})$, is computed, using the neighbors' distance to the current decoder's output representation, $d(\bm{f}(\bm{x},\bm{y}_{<t}), \cdot)$:
\begin{align}
    \label{eq:retrieval_distribution}
    &p_{k\mathrm{NN}}(y_t|\bm{y}_{<t}, \bm{x}) =\\ 
    &\dfrac{\sum_{(\bm{k}_j,v_j)\in \mathcal{N}}\mathbbm{1}_{y_t=v_j} \exp \left(-d\left(\bm{k}_j,\bm{f}(\bm{x},\bm{y}_{<t})\right)/T\right)}{\sum_{(\bm{k}_j,v_j)\in \mathcal{N}} \; \exp \left(-d\left(\bm{k}_j,\bm{f}(\bm{x},\bm{y}_{<t})\right)/T\right)}, \nonumber
\end{align}
where $T$ is the softmax temperature, $k_j$ denotes the key of the $j^{th}$ neighbor and $v_j$ its value. 
Finally, the two distributions, $p_{\mathrm{NMT}}(y_t|\bm{y}_{<t}, \bm{x})$ and $p_{k\mathrm{NN}}(y_t|\bm{y}_{<t}, \bm{x})$, are combined, by performing interpolation, to obtain the final distribution, which is used to generate the translation through beam search:
\begin{align}
    \label{eq:interpolation}
    p(y_t|\bm{y}_{<t}, \bm{x})= &\,\, (1-\lambda) \; p_{\mathrm{NMT}}(y_t|\bm{y}_{<t}, \bm{x}) \\
    &+ \lambda \; p_{k\mathrm{NN}}(y_t|\bm{y}_{<t},\bm{x}), \nonumber
\end{align}
where $\lambda\in [0,1]$ is a hyperparameter that controls the weight given to the two distributions.

\section{Chunk-based $k$NN-MT}

\begin{figure*}[ht]
    \centering
    \includegraphics[width=\textwidth]{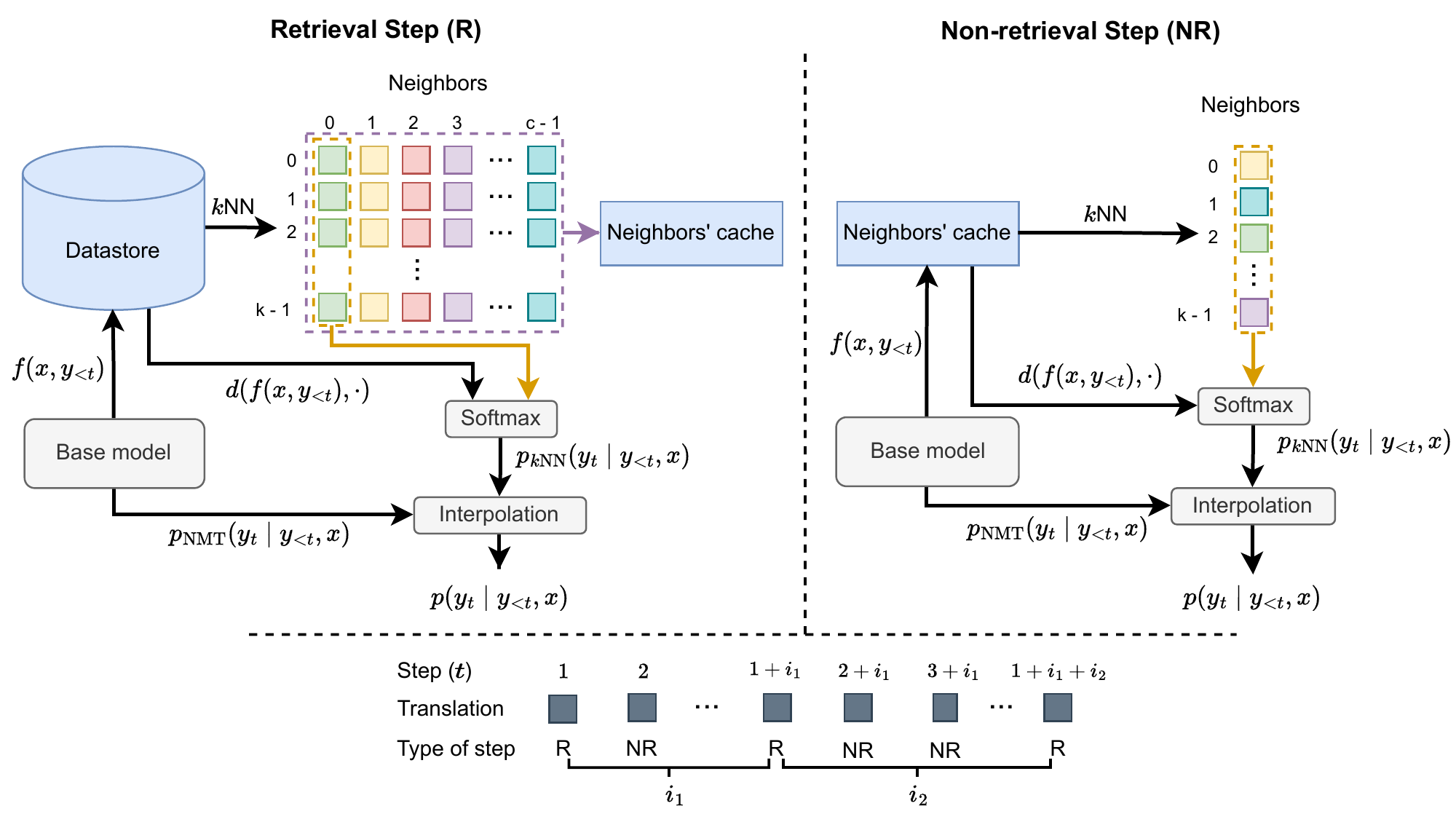}
    \caption{Chunk-based $k$NN-MT scheme. Top left: model procedure when retrieving neighbors from the datastore. Top right:  procedure when not performing retrieval. Bottom: retrieval schedule scheme. 
   } 
    \label{fig:scheme_chunk_knnmt}
\end{figure*}

We now describe our \textbf{chunk-based $k$NN-MT model}, illustrated in Figure \ref{fig:scheme_chunk_knnmt}. We first describe the  datastore creation (\S \ref{sec:building_datastore}) and how we can retrieve chunks of tokens from it (\S \ref{sec:retrieving_chunks}), describing how they are used by the model (\S \ref{sec:local_datastore}).  Finally, we describe how to select the steps in which the model performs retrieval (\S \ref{sec:selecting_steps}).

\subsection{Building the datastore}
\label{sec:building_datastore}
For the model to retrieve a chunk of tokens of size $c$ instead of a single token, we first need to build a datastore $\mathcal{D}$ which also consists of a key-value memory, where each entry key is the decoder's output representation $\bm{f}(\bm{x},\bm{y}_{<t})$, but the value is now a \textit{chunk} of target tokens $y_{t:(t+c-1)}$: \footnote{When the chunk size $c$ is larger than the number of the remaining tokens in the sentence, $N-(t-1)$, we add padding tokens to complete the chunk.} 
\begin{align}
    \mathcal{D} = &\{\left(\bm{f}(\bm{x},\bm{y}_{<t}\right),y_{t:(t+c-1)}) \; \forall \; t \mid (\bm{x},\bm{y}) \in \mathcal{S}\}.
\end{align}
%where $\mathcal{S}$ is a set of parallel sentences. 
Note that the chunks are sliding windows, \textit{i.e.}, they overlap.

\subsection{Retrieving chunks of tokens}
\label{sec:retrieving_chunks}
At inference time, when performing retrieval, the model searches the datastore and retrieves the set of $k$ nearest neighbors $\mathcal{N}$, for each beam hypothesis and example in the current batch. We now describe several strategies to manipulate the retrieved chunks of tokens during generation.

\subsubsection{Maintaining the chunk order}
Since in the original sentences used to build the datastore each chunk is composed of an ordered sequence of tokens, the simplest way for the model to use the retrieved tokens is to consider them in the same order. 
For this, we simply need to compute a retrieval distribution for each token in the chunk, always using the same retrieval distances but aligning the chunk tokens with the corresponding time step, \textit{i.e.} at the retrieval step we consider the first token of each neighbor chunk, at the following step we consider the second token, and so on.
We can compute this retrieval distribution as it is done for the $k$NN-MT, in Eq.~\ref{eq:retrieval_distribution}, just modifying the token indices according to the step.
However, by doing this, we are ignoring the remaining tokens in the chunk, which can also contain relevant information for the current prediction. Also, the tokens that are generated in the previous steps $t, \dots, t+j-1$ might not be well ``aligned'' with the next tokens in the $j$th position of the chunk for all neighbors.

\subsubsection{Neighbors' Cache}
\label{sec:local_datastore}
To avoid the limitation stated above, we propose using a neighbors' cache instead. We keep the tokens of the retrieved chunks in this cache, so that the model has a higher flexibility about which tokens to select for the current step: it has access to all the tokens present in the retrieved chunks. 
The neighbors' cache, $\mathcal{M}$, consists of a key-value memory, where a key is the decoder's output representation, $\bm{f}(\bm{x},\bm{y}_{<t})$, and a value is the corresponding target token $y_t$, as in the datastore:
\begin{align}
    \mathcal{M} = &\{\left(\bm{f}(\bm{x},\bm{y}_{<t+i}\right),y_{t+i}) \; \forall \; 0 \leq i < c \; \forall \; y_t  \mid \nonumber \\  & y_{t:(t+c-1)} \in \mathcal{N}\}.
\end{align}

Note, however, that this cache requires having the decoder state for every token in the chunk, not just for the first one. Therefore, we need to have this information available in the datastore:\footnote{To avoid having the same decoder states in several entries, we just store pointers to an array of decoder states.}
\begin{align}
     \mathcal{D} = \; & \{\left(\bm{f}(\bm{x},\bm{y}_{<t}\right),y_{t:(t+c-1)}, \nonumber \\ & [\bm{f}(\bm{x},\bm{y}_{<t}),\dots,\bm{f}(\bm{x},\bm{y}_{<t+c-1})]) \nonumber \\ & \forall \; t \mid   (\bm{x},\bm{y}) \in \mathcal{S}\}.
\end{align}
Then, as shown in the diagram of Figure \ref{fig:scheme_chunk_knnmt}, at the retrieval steps, the model first searches for the nearest neighbors in the datastore, then builds the neighbors' cache with the tokens of the retrieved chunks, then finally uses the first token of each chunk to compute the current retrieval distribution. 
In contrast, at the non-retrieval steps, the model searches for the nearest neighbors in the neighbors' cache instead of retrieving from the datastore. Then, to compute the retrieval distribution it also uses the softmax, as in Eq. \ref{eq:retrieval_distribution}, but with a different softmax temperature, $T'$. 
To incorporate the retrieved tokens, it performs interpolation as before, Eq. \ref{eq:interpolation}, replacing the hyperparameter $\lambda$ by $\lambda'$.\footnote{We use a different  temperature and interpolation coefficient at the non-retrieval steps because the number of entries in the neighbors' cache is much smaller than in the datastore, hence the neighbors might be less similar.}

\paragraph{Considering batch-beam-level neighbors. }
By building this neighbors' cache, the model can use all the tokens in the retrieved chunks that correspond to each beam hypothesis of the sentence being translated. 
However, it still ignores the chunks of tokens corresponding to the other beam hypotheses, which are often quite similar, and the other sentences being translated in the same batch, which can contain relevant contextual information if they belong to the same document. 

To also leverage these, we propose increasing the number of tokens that the model has access to, by combining the chunks retrieved for the different beam hypotheses and the different examples of the same batch. To do so, we simply need to build a single neighbors' cache for the current batch, to which we feed all the retrieved chunks' tokens.

\paragraph{Considering sentence-level neighbors. }
To also consider the chunks of tokens retrieved in previous steps of the generation of the current %batch of 
sentences, we propose to keep these in the neighbors' cache, instead of resetting the cache at each retrieval step.

\smallskip
\smallskip

We empirically compare these different proposed approaches in \S \ref{sec:incorporating}.

\subsection{Retrieval Steps Schedule}
\label{sec:selecting_steps}
As the need to perform retrieval slows down decoding considerably, having an efficient retrieval schedule is key to achieve decoding speedups. 
The simplest scheduling option corresponds to performing retrieval every $i$ steps.
However, we noticed empirically that it is beneficial to perform retrieval steps more frequently at the beginning of the sentence, as we will see in Table \ref{table:results_retrieval_steps} of \S \ref{sec:selecting}. To leverage this, we introduce the following schedule.

Having $k \in \{1,2, \ldots\}$ as the retrieval step's index and $t_k$ as the corresponding time step, (\textit{i.e.} $t_k$ is the position of the token generated after the $k\textsuperscript{th}$ retrieval step), we propose using a geometric progression to compute the interval (in tokens) between the $k\textsuperscript{th}$ and $(k+1)\textsuperscript{th}$ retrieval steps, $i_k = t_{k+1} - t_k$:
\begin{equation}
     \label{eq:exponential}
     i_k = \lfloor\min \left(i_{\max}, \: i_{\min} \times  2^{\,r\,t_k}\right)\rfloor,
\end{equation}
where $i_{\max}$ and $i_{\min}$ are hyperparameters that define the maximum and minimum interval between retrieval steps, the rate at which the interval increases is defined as $r={\frac{1}{2} i_{\max}}/{|\bm{x}|}$ where $|\bm{x}|$ is the source sentence size, and $\lfloor \cdot \rfloor$ denotes the floor function.\footnote{We always consider that the first retrieval step occurs at the first step of the generation ($t_0$=1).} By using this progression, the frequency with which the model performs retrieval decays along the generation, until the interval between retrieval steps reaches $i_{\max}$. For example, with $i_{\min}=2$, $i_{\max}=16$, and $|\bm{x}|=20$ the model performs retrieval at steps: $\{1,3,7,20,36,52,\dots\}$. 
Note that the chunk size, $c$, is independent of the interval between retrieval steps.

\begin{table*}[ht]
\centering \small
\setlength{\tabcolsep}{1.1ex}
\begin{tabular}
{lccccc@{\hspace{3.4ex}}ccccc}
\toprule
& \multicolumn{5}{c}{BLEU} & \multicolumn{5}{c}{COMET} \\
& Medical & Law & IT & Koran & Average & Medical & Law & IT & Koran & Average \\
\midrule
%\textbf{Parametric models} \\
Base MT & 40.01 & 45.64 & 37.91 & 16.35 & 34.98 & .4702 & .5770 & .3942 & -.0097 & .3579 \\
Fine-tuned & 50.47 & 56.56 & 43.82 & 21.54 & 43.10 & .5871 & .6906 & .5856 & .0484 & .4779 \\
\midrule
%\textbf{Semi-parametric models} \\
$k$NN-MT & 54.47 & 61.23 & 45.96 & 21.02 & 45.67 & .5760 & .6781 & .5163 & .0480 & .4546 \\
Efficient $k$NN-MT & 51.90 & 57.82 & 44.44 & 20.11 & 43.57 & .5513 & .6260 & .4909 & -.0052 & .4158 \\
Chunk-based $k$NN-MT (ours) & 53.16 & 59.65 & 44.18 & 19.33 & 44.08 & .5551 & .6257 & .4854 & .0039 & .4175  \\
\bottomrule
\end{tabular}
\caption{BLEU and COMET scores on the multi-domains test set, for a batch size of 8.}
\label{table:results}
\end{table*}

\begin{figure*}[ht]
    \centering
    \includegraphics[width=4.06cm]{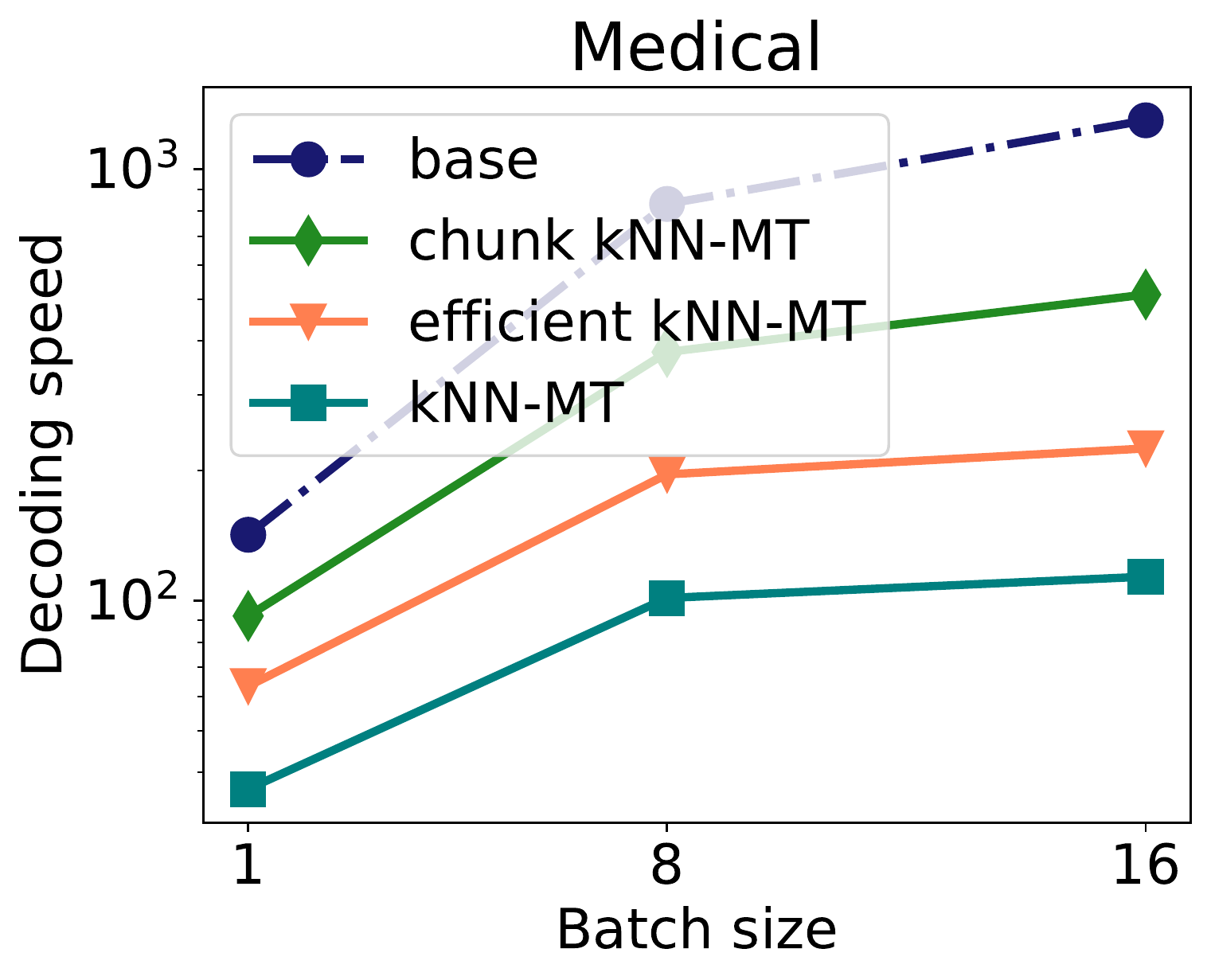}
    \includegraphics[width=3.88cm]{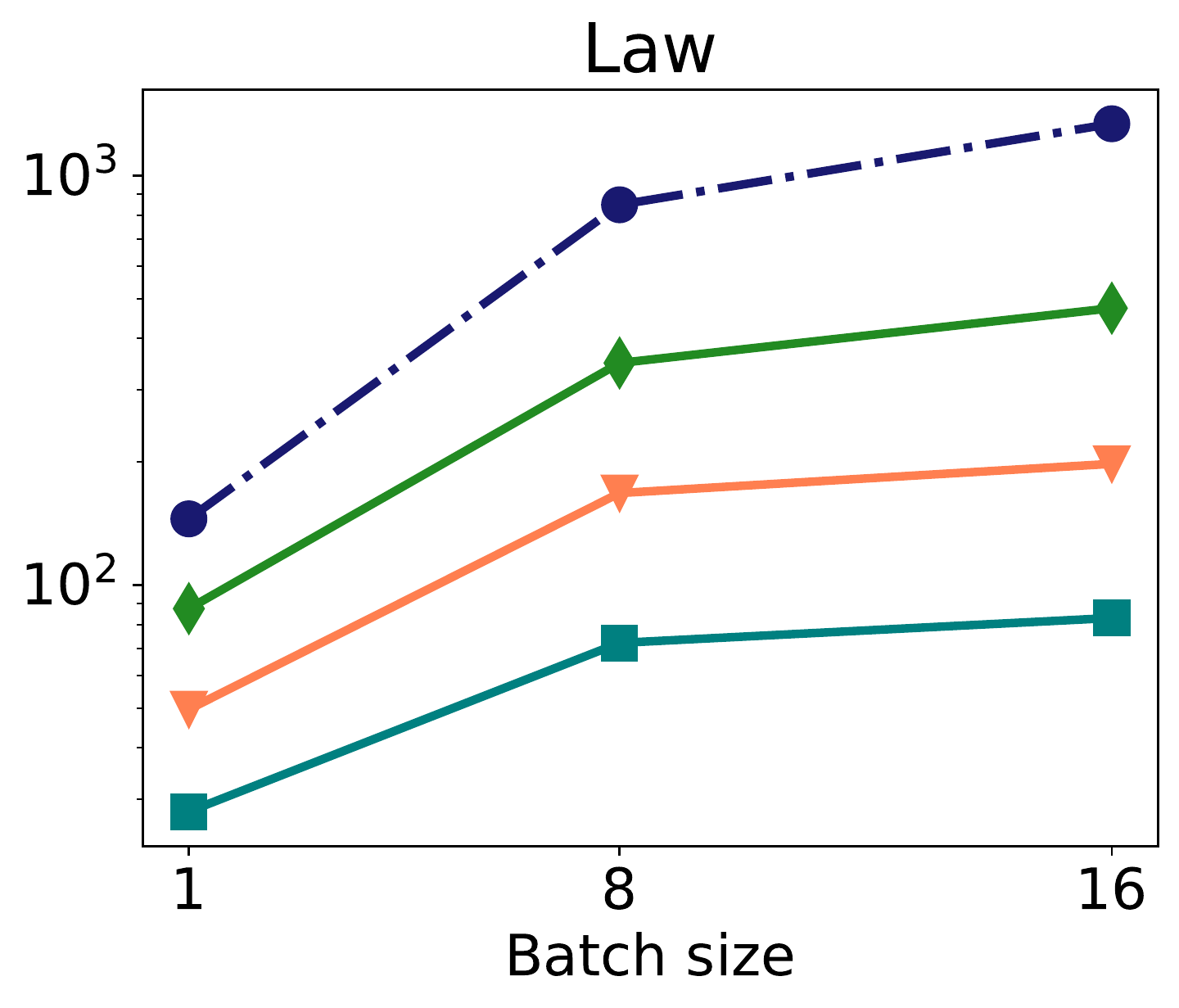}
    \includegraphics[width=3.88cm]{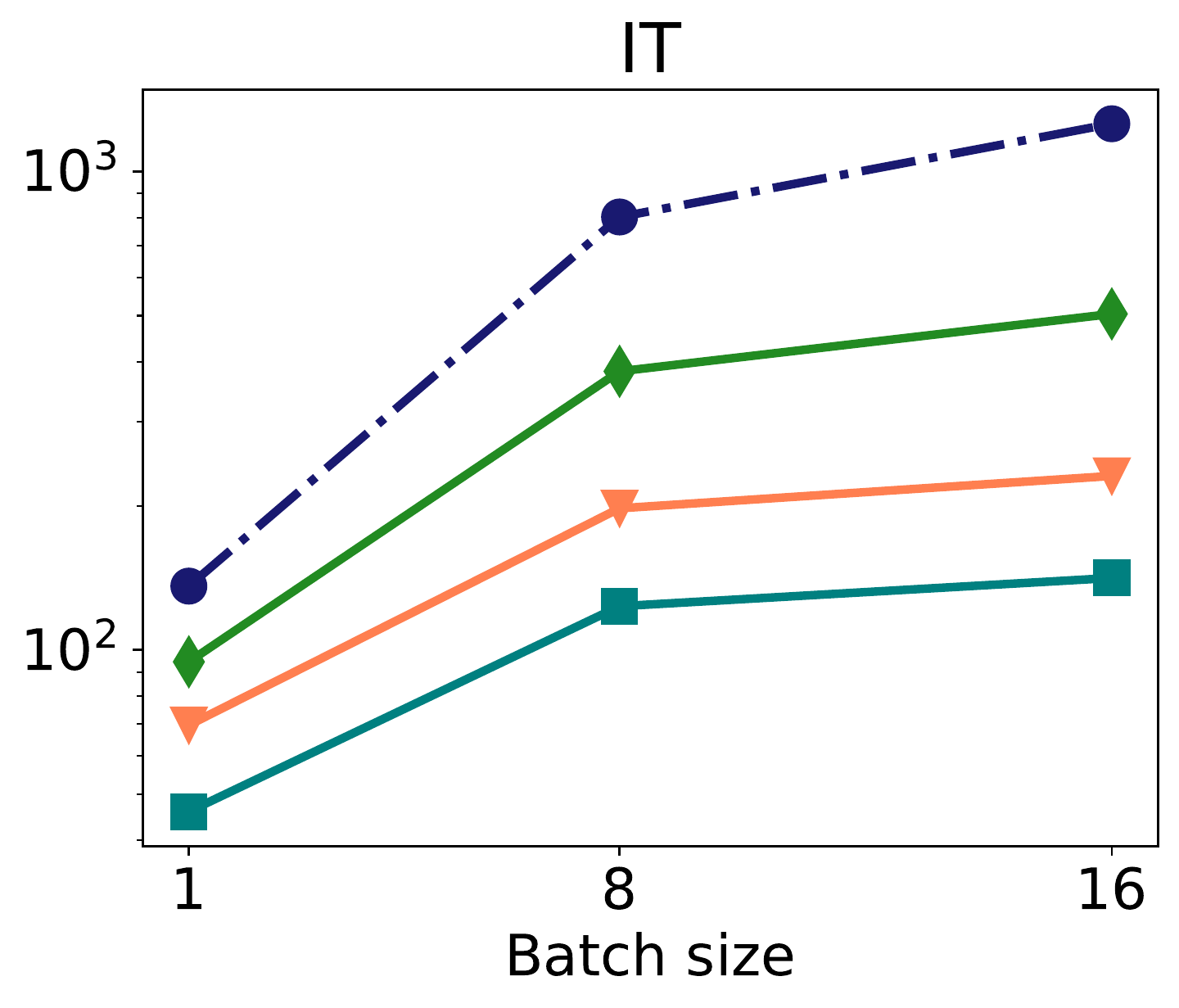}
    \includegraphics[width=3.88cm]{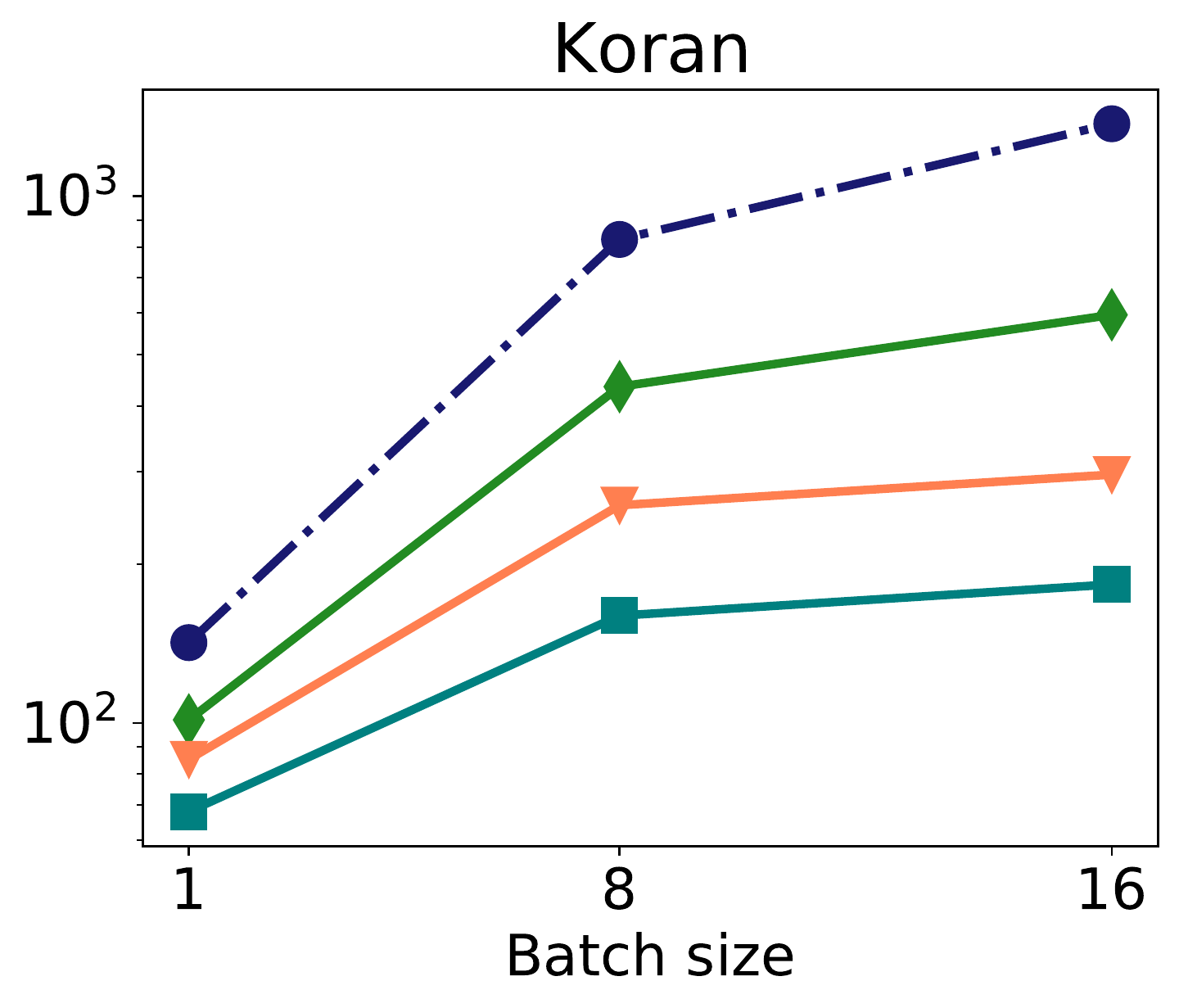}
    \caption{Plots of the decoding speed (tokens per second) for the different models on the medical, law, IT, and Koran domains, for different batch sizes (1,8,16). The generation speed (y-axis) is in log scale.}
    \label{fig:decoding_speed}
\end{figure*}

\section{Experiments}
To understand if the chunk-based $k$NN-MT is able to maintain the translation quality while speeding-up decoding, we performed experiments on domain adaptation (\S \ref{sec:domain_adaptation}) and on-the-fly adaptation (\S \ref{sec:on_the_fly_adaptation}).

\subsection{Domain Adaptation}
\label{sec:domain_adaptation}
\paragraph{Dataset and metrics. } For domain adaptation, we perform experiments on the Medical, Law, IT, and Koran domain data of the multi-domains dataset introduced by \citet{koehn2017six} using the splits redefined by \citet{aharoni2020unsupervised}. To build the datastores, we use the training sets that have 6,903,141, 19,061,382, 3,602,862, and 524,374 tokens, respectively. The validation and test sets have 2,000 examples for each domain. 
For evaluation, we use BLEU \citep{papineni2002bleu,post2018call} and COMET \citep{rei2020comet}. 

\paragraph{Models. }
As a baseline, we consider the fully parametric base MT model: the winning system from the WMT'19 German-English news translation task \citep{ng2019facebook} (with 269M parameters), available in Fairseq \cite{ott2019fairseq}. 
We also compare our chunk-based $k$NN-MT model with other models that have access to the domain-specific training data:  the base model above fine-tuned on the domain-specific datasets, the vanilla $k$NN-MT model \citep{khandelwal2020nearest}, and the Efficient $k$NN-MT model from \citet{martins2022efficient}. 

\paragraph{Settings. } 
For all models that perform retrieval, we retrieve $k=8$ neighbors and select the hyperparameter $\lambda$ for each method and each domain by performing grid search on $\lambda \in \{0.5,0.6,0.7,0.8\}$.  
For the chunk-based $k$NN-MT, we also perform grid search on $\lambda' \in \{0.4,0.5,0.6\}$ and $T' \in \{1,2,3\}$.
The selected values for the hyperparameters for each model for each domain, validated on the validation set, are stated in Table \ref{table:hyperparameters} of App. \ref{sec:hyperparameters}.
We use the softmax temperatures proposed by \citet{khandelwal2020nearest} and for the efficient $k$NN-MT, we use the efficiency methods' parameters proposed by \citet{martins2022efficient}. Where not stated otherwise, we use the chunk-based $k$NN-MT with chunks of size $c=16$, with a sentence-level neighbors' cache, and use the geometric progression heuristic, in Eq. \ref{eq:exponential}, to select the retrieval steps, with $i_{\min} =2$ and $i_{\max} =16$, since this is the setting that leads to the best trade-off between translation quality and decoding speed on the validation set. 
We also follow \citet{martins2022efficient} and use PCA to reduce the datastore keys' dimension to 256 and the neighbors' cache keys' size to 64.
To perform search in the datastore and in the neighbors' cache, we use FAISS \citep{johnson2019billion}.

For the fine-tuned model, we perform fine-tuning for a maximum of 20 epochs on each domain. We perform grid search on the validation set, using different learning rates, $\eta  \in \{5\times 10^{-6}, 1\times 10^{-5}, 5\times 10^{-5}, 1\times 10^{-4} \}$ and two different learning rate schedules (reducing learning rate on plateau and by the inverse square root) with and without warmup during 1 epoch. The selected hyperparameters are stated in Table \ref{table:hyperparameters_finetuned} of App. \ref{sec:hyperparameters}.

\paragraph{Computational infrastructure. }
All experiments were performed on a server with 3 RTX 2080 Ti (11 GB), 12 AMD Ryzen 2920X CPUs (24 cores), and 128 Gb of RAM. For the decoding speed measurements, we ran each model on a single GPU while no other process was running on the server, to have a controlled environment. The nearest neighbor search in the datastore is performed on the CPUs, since not all datastores fit into GPU memory.

\subsubsection{Results}
\label{sec:results}
The translation scores are reported in Table \ref{table:results}. We can see that the decrease of translation quality when comparing the chunk-based $k$NN-MT model with the vanilla $k$NN-MT model is not substantial in terms of BLEU (-1.5 points on average) and COMET (-.04 points on average). It can also be seen that the chunk-based $k$NN-MT model leads to considerably better translation scores than the base MT model (+9.1 BLEU and +.06 COMET points on average) and to slightly better results than the efficient $k$NN-MT model in terms of BLEU.
When comparing fine-tuning the base model with the use of semi-parametric models, the results are not conclusive: in terms of BLEU, the semi-parametric models lead to better translations, but according to COMET this is not the case.
We present translation examples for the different domains in App. \ref{sec:examples}.

\subsubsection{Decoding speed}
As can be seen in Figure \ref{fig:decoding_speed}, the chunk-based $k$NN-MT model leads to a decoding speed up to two times higher than the decoding speed of the efficient $k$NN-MT model of \citet{martins2022efficient} and up to four times higher than that of the vanilla $k$NN-MT model of \citet{khandelwal2020nearest}. The chunk-based $k$NN-MT model is also able to reduce the decoding speed gap to the base MT model to a factor of two, compared to a factor of four from previous work. Moreover, according to the results on Table \ref{table:results} this speed-up comes without substantially harming the model's translation quality.

\subsubsection{What is the best way to incorporate the retrieved tokens?}
\label{sec:incorporating}
To understand which chunk incorporation strategy works best, we perform a comparison using chunks of size $c=6$ and performed retrieval every 6 steps ($i=6$).
\begin{table}[t]
\centering \small
\setlength{\tabcolsep}{.55ex}
\begin{tabular}
{lccccc}
\toprule
& Medical & Law & IT & Koran & Average \\
\midrule
Maintain order & 45.17 & 51.48 & 40.11 & 17.87 & 38.66 \\
\midrule 
\textbf{Neighbors' Cache} \\
Single chunk     & 48.01 & 53.81 & 42.09 & 18.49 & 40.60 \\
Beam-batch-level & 51.77 & \textbf{58.80} & 42.91 & 19.46 & 43.24 \\
Sentence-level   & \textbf{51.86} & 58.68 & \textbf{43.44} & \textbf{19.79} & \textbf{43.44} \\
\bottomrule
\end{tabular}
\caption{BLEU scores on the multi-domains test set, for a batch size of 8, with $c=6$ and $i=6$.}
\label{table:results_local_memory}
\end{table}
The results reported in Table \ref{table:results_local_memory} show that using a neighbors' cache leads to substantially better BLEU scores. 
We can also see that having a beam-batch-level cache improves the BLEU score and that keeping the tokens from the previous retrieved chunks in the neighbors' cache further improves the translation quality.

\subsubsection{When to perform retrieval?}
\label{sec:selecting}
\begin{table}[t]
\centering \small
\setlength{\tabcolsep}{.7ex}
\begin{tabular}
{lccccc}
\toprule
& Medical & Law & IT & Koran & Average \\
\midrule
$i=6$                 & 51.86 & 58.68 & 43.44 & \textbf{19.79} & 43.44 \\
$i=8$                 & 51.36 & 58.28 & 43.04 & 19.25 & 42.98 \\
GP ($i_{\max} = 8$)  & \textbf{53.38} & \textbf{59.87} & \textbf{44.41} & \textbf{19.74} & \textbf{44.35} \\
GP ($i_{\max} = 16$) & 53.16 & 59.65 & 44.18 & 19.33 & 44.08 \\
GP ($i_{\max} = 32$) & 52.81 & 58.96 & 43.51 & 19.22 & 43.63 \\
\bottomrule
\end{tabular}
\caption{BLEU scores on the multi-domains test set, for a batch size of 8. When using the geometric progression heuristic (GP) the average interval with $i_{\max} = 16$ is 5.97 and with $i_{\max} = 32$ is 6.85.}
\label{table:results_retrieval_steps}
\end{table}

\begin{table}[t]
\centering \small
\setlength{\tabcolsep}{.9ex}
\begin{tabular}
{lccccc}
\toprule
& Medical & Law & IT & Koran & Average \\
\midrule
$i=6$                & 374 & 328 & 398 & 441 & 385 \\
$i=8$                & 421 & 374 & \textbf{429} & 470 & 424 \\
GP ($i_{\max} = 8$)  & 354 & 307 & 358 & 405 & 356 \\
GP ($i_{\max} = 16$) & 397 & 368 & 393 & 445 & 401 \\
GP ($i_{\max} = 32$) & \textbf{436} & \textbf{423} & 418 & \textbf{481} & \textbf{440} \\
\bottomrule
\end{tabular}
\caption{Decoding speed (tokens per second) on the multi-domains test set, for a batch size of 8, with $c=i_{\max}$. }
\label{table:speed_retrieval_steps}
\end{table}
To understand how we should select the retrieval steps, we compare performing retrieval every $i=6$ or $i=8$ steps against using the proposed geometric progression (GP), Eq. \ref{eq:exponential}, with $i_{\min} = 2$ and $i_{\max} = 8$, $i_{\max} = 16$, or $i_{\max} = 32$, to compute the interval between retrieval steps. For this comparison, we used the model with a sentence-level neighbors' cache and considered $c=i$ or $c=i_{\max}$ if using the geometric progression. 
We report the BLEU scores in Table \ref{table:results_retrieval_steps} and the corresponding decoding speeds in Table \ref{table:speed_retrieval_steps}.
This comparison shows that performing retrieval steps more frequently at the beginning of the translation, by using the the proposed geometric progression heuristic (GP), leads to better BLEU scores while having a higher decoding speed.

\subsection{On-the-fly Adaptation}
\label{sec:on_the_fly_adaptation}
To understand how the chunk-based $k$NN-MT model 
behaves in a realistic scenario where the data arrives in streams, we  performed experiments where the model is continuously adapted on the fly.

\begin{figure}[h!]
    \centering
    \includegraphics[width=5.1cm]{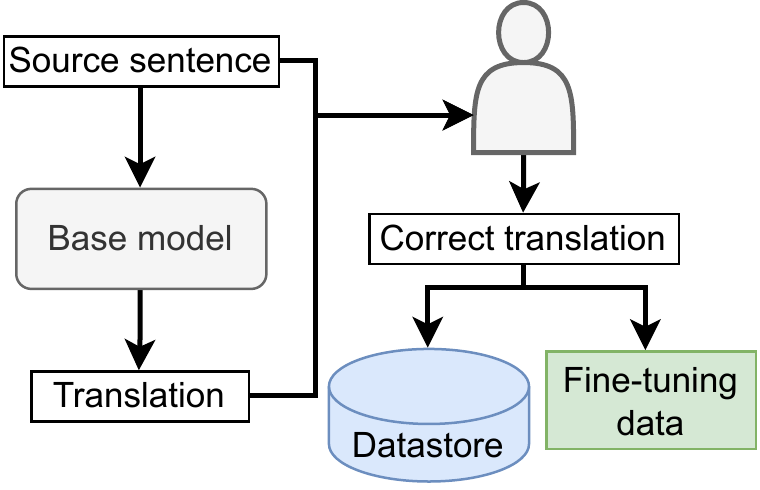}
    \caption{On-the-fly adaptation scheme. After the model translates an example, a human translator (e.g., a post-editor) corrects the generated translation which is added to the fine-tuning data or the domain-specific datastore. We use the references to simulate  human translations.}
    \label{fig:on_the_fly_adaptation_scheme}
\end{figure}

\begin{figure*}[t]
    \centering
    \includegraphics[width=6.3cm]{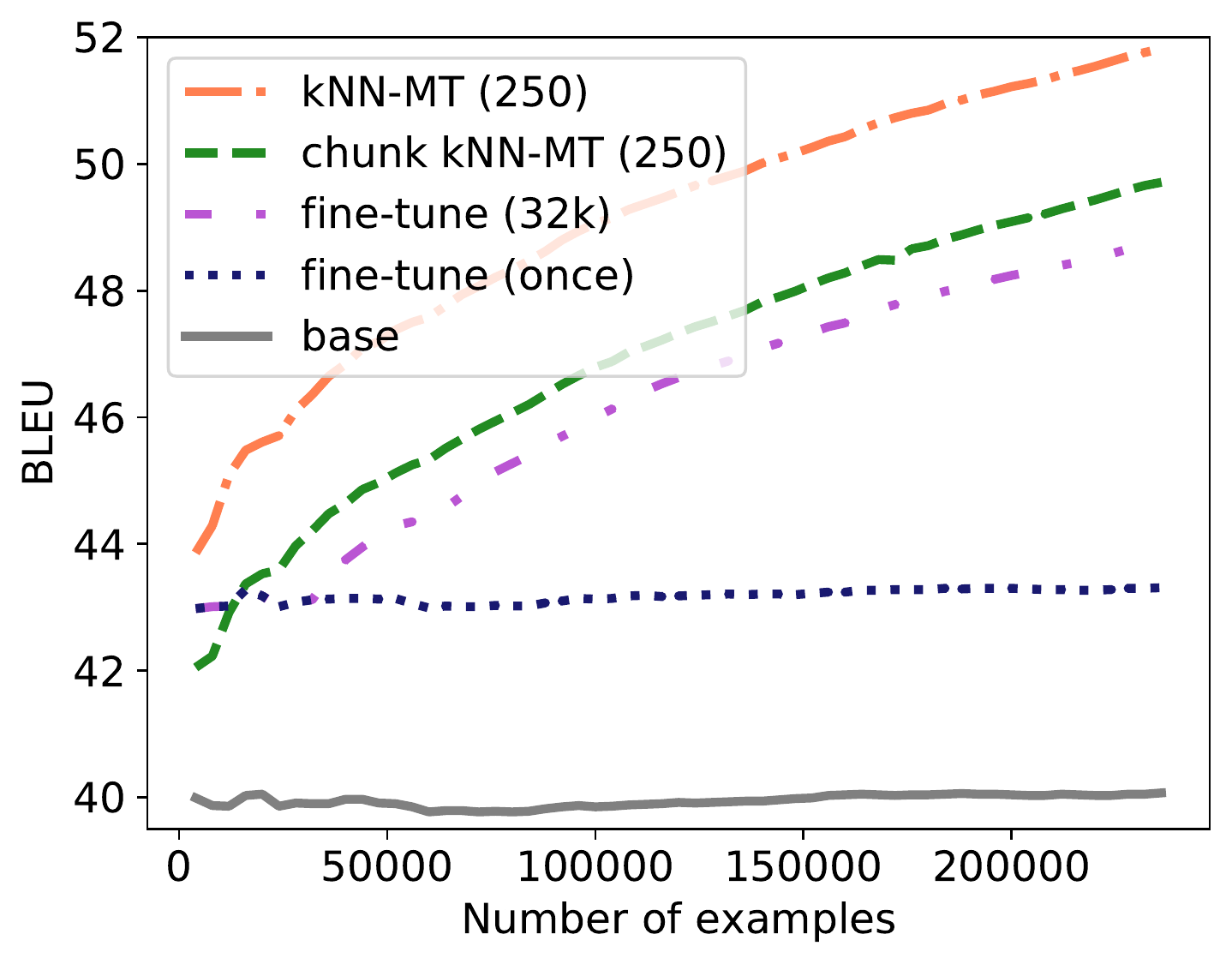} \qquad
    \includegraphics[width=6.4cm]{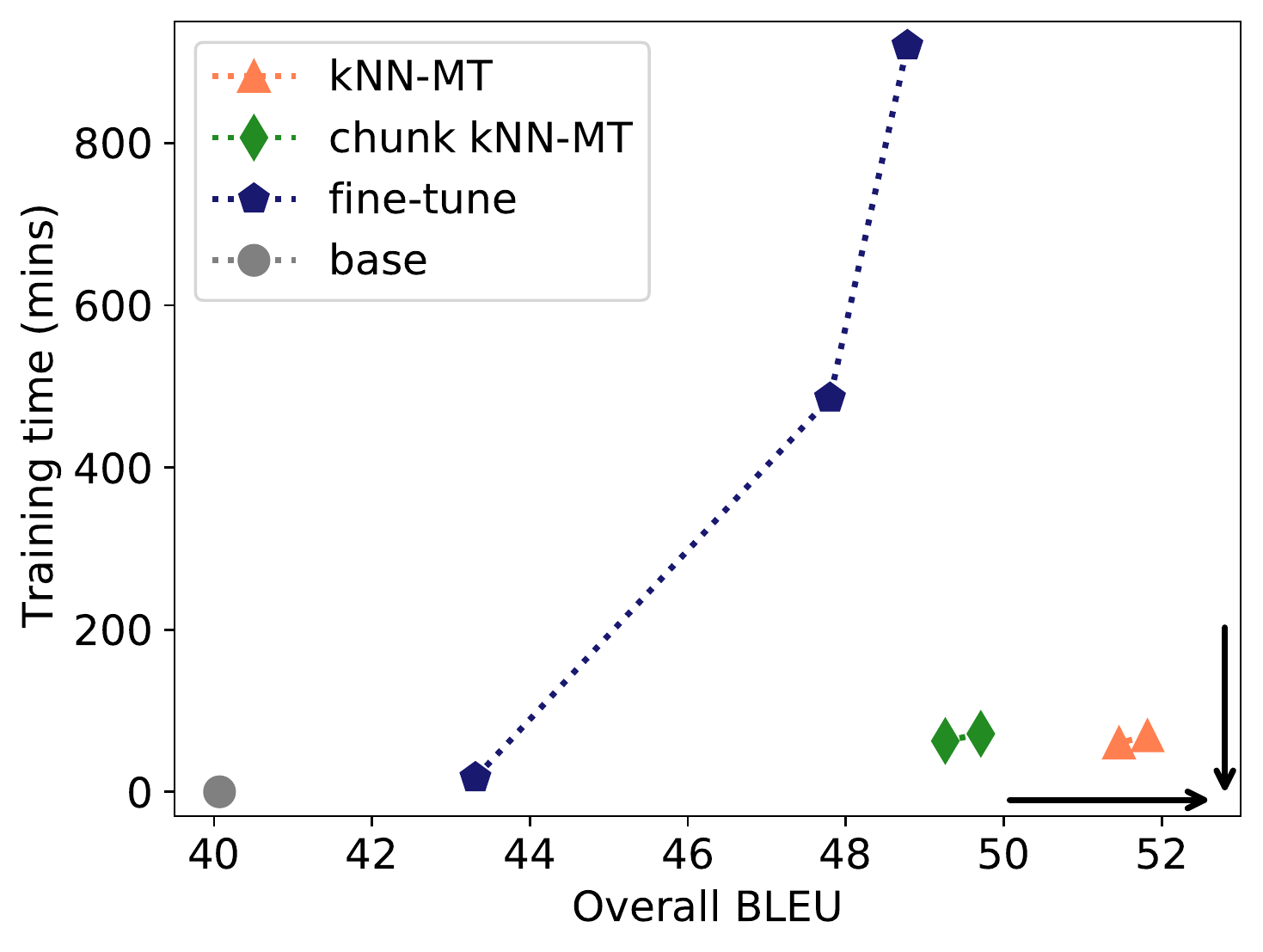}
    \includegraphics[width=6.4cm]{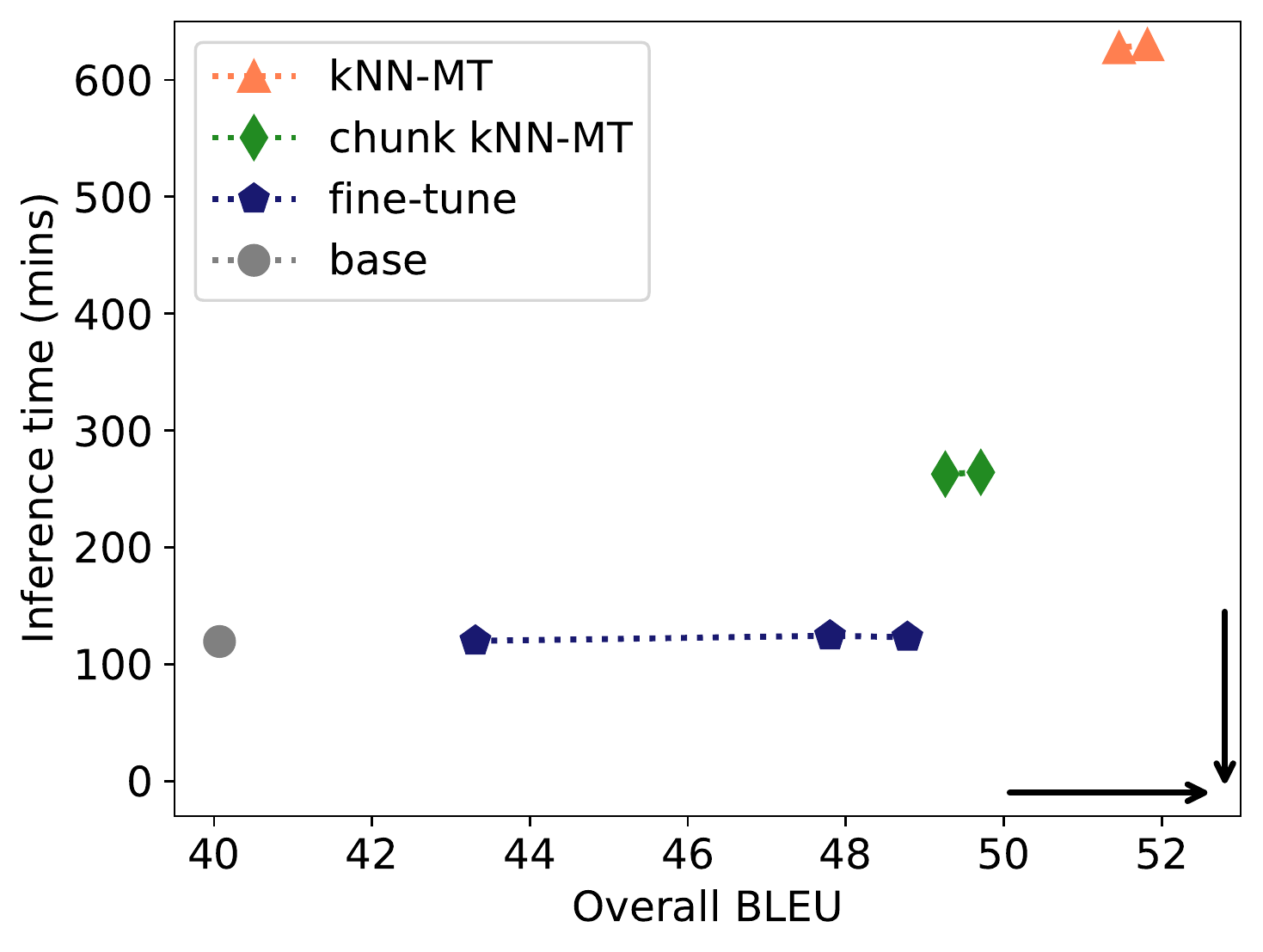} \qquad
    \includegraphics[width=6.4cm]{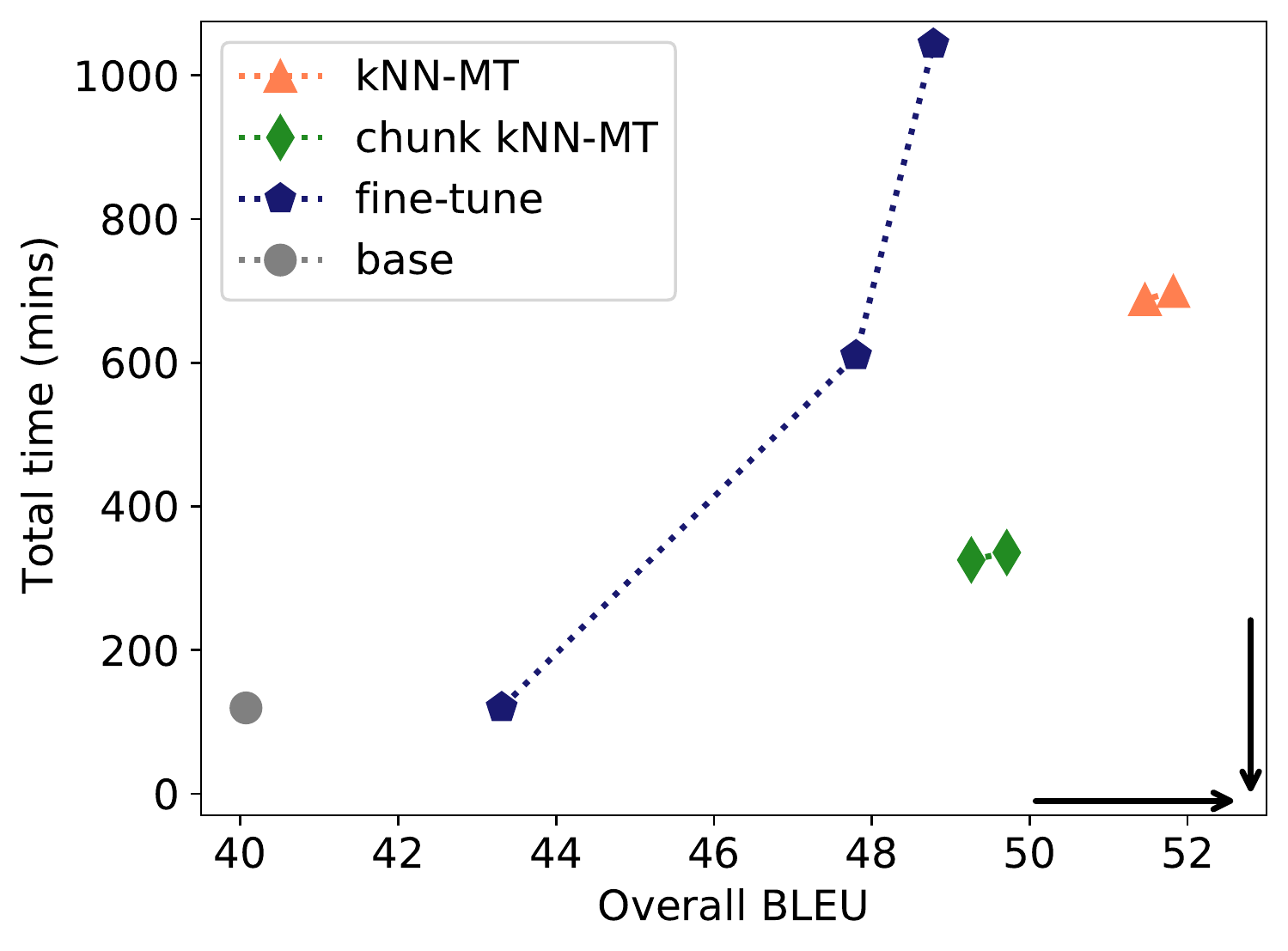}
    \caption{Analysis of the on-the-fly adaptation experiments on the medical domain. Top left: BLEU scores measured every 4,000 examples. Top right: Time (in minutes) spent by each model on training / creating and updating the datastore and the corresponding BLEU score. Bottom left: Inference time (in minutes) spent by each model to translate the whole set of examples and its BLEU score. Bottom right: Total time ($\text{training} + \text{inference}$) spent by each model to translate the set of examples and its BLEU score. For the time plots, the lower right corner is better.}
    \label{fig:results_on_the_fly_adaptation}
\end{figure*}

\paragraph{Task description. } 
In this task, we attempt to simulate a real scenario, described in Figure \ref{fig:on_the_fly_adaptation_scheme}, in which we have a model that translates sentences and a human translator (e.g., a post-editor) who corrects the generated translations. Our goal is to understand whether it is better to use the corrected translations to repeatedly fine-tune the base model or to add the corrected translations to the datastore without touching the base model. 
To do so, we consider that, before starting translation, we have access to 10\% of the dataset (we use the training sets of the medical and law domains) and the goal is to translate the remaining 90\% examples. To simulate the existence of human translators who correct the %automatically 
generated translations, we consider that, after the model translates a block of %source 
sentences, it has access to the corresponding reference sentences.

\paragraph{Models. } As baselines, we use the base MT model and the base model fine-tuned on the initially available 10\% of data (fine-tune (once)). We also compare fine-tuning the base model on the initially available data and then finetuning after every 32,000 and every 64,000 examples, with all the available data at the time. 
In all cases, we fine-tuned the model for a maximum of 5 epochs with a learning rate of $5\times 10^{-5}$ using the inverse square root scheduler. 
Concerning the semi-parametric models, we compare the $k$NN-MT model and the chunk-based $k$NN-MT model, building the initial datastore with the initially available data and adding new examples to the datastore after every 250 or 1,000 sentences.
We used the same configurations for the chunk-based $k$NN-MT model, as in \S \ref{sec:domain_adaptation}: chunks of size $c=16$, sentence-level neighbors' cache, and the geometric progression heuristic to select when to perform retrieval, with $i_{\min} =2$ and $i_{\max} =16$. We also use the same hyperparameters, stated on Table \ref{table:hyperparameters} of App. \ref{sec:hyperparameters}.

\subsubsection{Results}
Figure \ref{fig:results_on_the_fly_adaptation} contains the results of the on-the-fly adaptation experiments on the medical domain. On the top left plot, we see that both the $k$NN-MT and the chunk-based $k$NN-MT lead to higher BLEU scores than the fine-tuned models. 
Also, on the top right, we see that the time needed to add examples to the datastore is much shorter than the time needed to fine-tune the model. This comes at the cost of a higher inference time (bottom left). However, we can see that the chunk-based $k$NN-MT model is able to substantially reduce the inference time gap to the fully-parametric models. Also, the bottom right plot shows that the chunk-based $k$NN-MT has a shorter total time than $k$NN-MT and the models fine-tuned every 32,000 and 64,000 steps.
Concerning the fine-tuned models, there is a BLEU increase when the model is fine-tuned more often. However, this leads to a substantially higher training time. On the other hand, increasing the datastore updates frequency leads to small BLEU improvements and small increases in training time. 
The results on the law domain show similar results (App. \ref{sec:on_the_fly_adaptation_law}).

\section{Related Work}

\paragraph{Semi-parametric models. }
Semi-parametric models, which augment a parametric model with a retrieval component, 
have been shown to be effective on several text generation tasks. For language modeling, \citet{khandelwal2019generalization} proposed the $k$-nearest neighbor language ($k$NN-LM), in which a language model is augmented with token-based retrieval and uses probability interpolation to incorporate these tokens. \citet{yogatama2021adaptive} proposed to integrate the retrieved tokens with a gating mechanism. \citet{borgeaud2021improving} proposed to retrieve chunks of tokens and incorporate them with cross-attention, using datastores with trillions of tokens. 
To increase the $k$NN-LM's decoding speed, \citet{he2021efficient} proposed a range of techniques, such as datastore pruning, dimension reduction, and adaptive retrieval. \citet{alon2022neuro} proposed adding pointers to the next token on the original corpus to the datastore entries, so that the model can consider the pointed entries instead of performing retrieval. 
Similarly to our approach, this saves retrieval steps by leveraging the original corpus sequences, but in our case, we  do not limit the candidate tokens to be the following ones and consider the succeeding tokens even if the model has not generated the same prefix token(s).

For machine translation, \citet{gu2018search} introduced a semi-parametric model which uses an out-of-the-box search engine to retrieve similar sentence pairs, and incorporate them with shallow and deep fusion. \citet{zhang2018guiding} proposed to retrieve $n$-grams and use them to up-weight token probabilities. \citet{bapna2019non} proposed to retrieve sentences similar to the source's $n$-grams, and incorporate them with attention. 
More recently, \citet{khandelwal2020nearest} proposed the $k$NN-MT model %, an adaptation of the $k$NN-LM, 
which \citet{zheng2021adaptive} extended with a network that determines the number of retrieved tokens to consider  and \citet{zheng2021non} proposed building the datastore using monolingual sentences.
As $k$NN-MT can be up to two orders of magnitude slower than a fully-parametric model, methods that improve its efficiency have been proposed.
\citet{meng2021fast} and \citet{wang2021faster} proposed the Fast and Faster $k$NN-MT, in which the model has a higher decoding speed, by creating a different datastore, based on the source sentence, for each example. \citet{martins2022efficient} propose efficient $k$NN-MT, which we use as baseline (\S \ref{sec:domain_adaptation}), by adapting the methods introduced by \citet{he2021efficient} to machine translation and introducing a retrieval distributions cache to speed-up decoding. In this paper, we show that the chunk-based $k$NN-MT model can further speed-up decoding, by retrieving chunks of tokens instead of a single token.

Semi-parametric models have also been applied to other tasks as question answering \citep{lewis2020retrieval,izacard2021leveraging,izacard2021distilling} and  dialogue generation \citep{weston2014memory,fan2021augmenting}.

\paragraph{Domain adaptation for machine translation. }
Domain adaptation consists of adapting generic models to domain-specific data. The most common method for domain adaptation in machine translation is fine-tuning the model on each domain, but this can be expensive and often leads to catastrophic forgetting \citep{saunders2021domain}. To simplify this, some work has proposed fine-tuning only part of the model \citep{wuebker2018compact,bapna2019simple,lin2021learning,liang2021finding}. \citet{farajian2017multi} performed on-the-fly adaptation, by fine-tuning the model on a set of retrieved examples for each source sentence. However, this still requires fine-tuning model parameters.

Several works have introduced  domain adaptation methods without the need to fine-tune the model.
\citet{eidelman2012topic}, \citet{hasler2014dynamic}, and \citet{su2015context} proposed using topic models while \citet{bertoldi2014online} proposed leveraging post-editing information.
More recently,  \citet{khandelwal2020nearest} proposed using semi-parametric models which retrieve from domain-specific datastores. The aim of our chunk-based $k$NN-MT method is to speed up $k$NN-MT's decoding, while maintaining its high translation quality.

\section{Conclusions}
In this paper, we propose a chunk-based $k$NN-MT model, which retrieves chunks of tokens from a datastore, instead of a single token. To do so, we proposed several alternatives to explore the retrieved chunks' tokens: keeping the original order or building a neighbors' cache. We also analyzed two approaches to select the retrieval steps: every $i$ steps or using a geometric progression heuristic to define the interval between retrieval steps.
Through experiments on domain adaptation, we showed that chunk-based $k$NN-MT leads to a considerable speed-up without substantially compromising the translation quality. Experiments on on-the-fly adaptation showed that chunk-based $k$NN-MT leads to high quality translations while being more efficient than previously proposed methods.

\section*{Limitations}
The scope of this paper is limited to the usage of small to medium size datastores, due to the memory requirements needed for big size datastores, for which the proposed model could be even more beneficial. 
%In this work, we performed experiments on German-English, using four datasets of different domains. 
%Thus, while the quality-speed trade offs investigated by our paper seem to depend more on the availability of domain-specific data than on the choice of language pair, further work is necessary to validate our approach to other language pairs and datasets.
Additionally, we use the decoding speed (tokens per second), training time (in minutes), and inference time (in minutes) to compare the efficiency of the different models. However, these metrics depend on the computational infrastructure used, and, consequently, the speed-up gains can vary when using different hardware.

\section*{Acknowledgments}
This work was supported by the European Research Council (ERC StG DeepSPIN 758969), 
by EU's Horizon Europe Research and Innovation Actions 
(UTTER, contract 101070631), 
by the P2020 project MAIA (contract 045909), by the Funda\c{c}\~ao para a Ci\^encia e Tecnologia through project PTDC/CCI-INF/4703/2021 (PRELUNA), contract UIDB/50008/2020, and by contract PD/BD/150633/2020 in the scope of the  Doctoral Program  FCT - PD/00140/2013 NETSyS. We thank Chris Dyer, Lei Yu, the SARDINE team members, and the reviewers for helpful discussion and feedback.

\bibliography{anthology,custom}
\bibliographystyle{acl_natbib}

\clearpage

\appendix

\section{Varying the Chunk Size along the Generation}
When using the geometric progression to compute the interval between retrieval steps  (\S \ref{sec:selecting_steps}), the model performs retrieval more frequently at the beginning of the generation of the translation. Because of this, we compare having a fixed chunk size equal to the maximum interval between retrieval steps ($c=i_{\max}$) with having the chunk size vary along the generation ($c_k=i_k$).
For this comparison, we compute the retrieval steps using the geometric progression with  $i_{\min} = 2$ and $i_{\max} = 16$, and use a sentence-level cache.
\begin{table}[h!]
\centering \small
\setlength{\tabcolsep}{1.2ex}
\begin{tabular}
{lccccc}
\toprule
& Medical & Law & IT & Koran & Average \\
\midrule
$c=i_{\max}$ & \textbf{53.16} & \textbf{59.65} & \textbf{44.18} & \textbf{19.33} & \textbf{44.08} \\
$c_k=i_k$  & 52.70 & 59.40 & 43.96 & 19.10 & 43.79 \\
\bottomrule
\end{tabular}
\caption{BLEU scores on the multi-domains test set, for a batch size of 8.}
\label{table:varying_chunk_size}
\end{table}

The results, in Table \ref{table:varying_chunk_size}, indicate that keeping the chunk size fixed leads to slightly better translation quality.

\section{Using Different Values for $k$.}
In order to understand how the number of retrieved chunks ($k$) affects the translation quality and the decoding speed, we compare using different values of $k$. For this comparison, we compute the retrieval steps using the geometric progression with  $i_{\min} = 2$ and $i_{\max} = 16$, and use a sentence-level cache.

\begin{table}[h]
\centering \small
\setlength{\tabcolsep}{1.3ex}
\begin{tabular}
{lccccc}
\toprule
& Medical & Law & IT & Koran & Average \\
\midrule
$k=2$  & 50.55 & 56.30 & 41.42 & 18.32 & 41.65 \\
$k=4$  & 51.46 & 58.28 & 43.55 & 19.07 & 43.09 \\
$k=8$  & 53.16 & 59.65 & 44.18 & 19.33 & 44.08 \\
$k=16$ & \textbf{53.75} & \textbf{60.64} & \textbf{45.37} & \textbf{19.99} & \textbf{44.94}\\
\bottomrule
\end{tabular}
\caption{BLEU scores on the multi-domains test set, for a batch size of 8.}
\label{table:results_k}
\end{table}

\begin{table}[h]
\centering \small
\setlength{\tabcolsep}{1.4ex}
\begin{tabular}
{lccccc}
\toprule
& Medical & Law & IT & Koran & Average \\
\midrule
$k=2$  & \textbf{452} & \textbf{392} & \textbf{452} & \textbf{494} & \textbf{448} \\
$k=4$  & 433 & 386 & 428 & 486 & 433 \\
$k=8$  & 397 & 368 & 393 & 445 & 401 \\
$k=16$ & 343 & 323 & 329 & 380 & 344 \\
\bottomrule
\end{tabular}
\caption{Decoding speed (tokens per second) on the multi-domains test set, for a batch size of 8. }
\label{table:speed_k}
\end{table}

We report the BLEU score and decoding speed for different values of $k$ in Tables \ref{table:results_k} and \ref{table:speed_k}, respectively. These results show that there is a trade-off between the translation quality and the decoding speed, when varying the number of retrieved neighbors ($k$).

\section{Experiments on ES-FR and ET-IT.}
To understand if the proposed model, chunk-based $k$NN-MT, performs well on language pairs and datasets other than the ones used in the main experiments (\S \ref{sec:results}), we perform experiments on Spanish-French (es-fr) and Estonian-Italian (et-it) on two datasets: EMEA and JRC-acquis \citep{TIEDEMANN12.463}. For this experiment, we used the multilingual model mBART50 \citep{tang2020multilingual}, compute the retrieval steps using the geometric progression with  $i_{\min} = 2$ and $i_{\max} = 16$, and use a sentence-level cache.

\begin{table}[h]
\centering \small
\setlength{\tabcolsep}{1.5ex}
\begin{tabular}
{lccccc}
\toprule
& \multicolumn{2}{c}{EMEA} & \multicolumn{2}{c}{JRC} \\
& es-fr & et-it & es-fr & et-it \\
\midrule
Base MT              & 5.02  & 16.44 & 19.27 & 17.94 \\
$k$NN-MT             & 36.65 & 38.79 & 45.87 & 39.88 \\
Chunk-based $k$NN-MT & 25.69 & 32.80 & 29.17 & 28.16 \\
\bottomrule
\end{tabular}
\caption{BLEU scores on the EMEA and JRC test sets, for a batch size of 8.}
\label{table:results_emea_jrc}
\end{table}

\begin{table}[h]
\centering \small
\setlength{\tabcolsep}{1.5ex}
\begin{tabular}
{lccccc}
\toprule
& \multicolumn{2}{c}{EMEA} & \multicolumn{2}{c}{JRC} \\
& es-fr & et-it & es-fr & et-it \\
\midrule
Base MT              & 189 & 214 & 189 & 205 \\
$k$NN-MT             & 37  & 37  & 26  & 27  \\
Chunk-based $k$NN-MT & 98  & 96  & 95  & 101 \\
\bottomrule
\end{tabular}
\caption{Decoding speed (tokens per second) on the EMEA and JRC test sets, for a batch size of 8. }
\label{table:speed_emea_jrc}
\end{table}

We report the BLEU scores and the decoding speeds (in tokens per second) on Tables \ref{table:results_emea_jrc} and \ref{table:speed_emea_jrc}, respectively. As can be seen, the chunk-based $k$NN-MT model is able to improve the translation quality considerably, when comparing with the base MT model, while leading to a decoding speed around 3 times faster than the vanilla $k$NN-MT model.

\section{Hyperparameters}
\label{sec:hyperparameters}

On Table \ref{table:hyperparameters} we report the values for the hyperparameters of the semi-parametric models: the interpolation coefficients $\lambda \in \{0.5,0.6,0.7,0.8\}$ and $\lambda' \in \{0.4,0.5,0.6\}$, and the retrieval softmax temperatures $T$ and $T' \in \{1,2,3\}$. For decoding we use beam search with a beam size of $5$.

On Table \ref{table:hyperparameters_finetuned} we report the values of the hyperparameters used to fine-tune the base model on each domain: learning rate, learning rate scheduler, and whether warmup steps were used.

\begin{table*}[hptb!]
\centering \small
\setlength{\tabcolsep}{1.25ex}
\begin{tabular}
{lcccc@{\hspace{5ex}}cccc@{\hspace{5ex}}cccc@{\hspace{5ex}}cccc}
\toprule
& \multicolumn{4}{c}{Medical} & \multicolumn{4}{c}{Law} & \multicolumn{4}{c}{IT} & \multicolumn{4}{c}{Koran} \\
& $\lambda$ & $T$ & $\lambda'$ & $T'$ & $\lambda$ & $T$ & $\lambda'$ & $T'$ & $\lambda$ & $T$ & $\lambda'$ & $T'$ & $\lambda$ & $T$ & $\lambda'$ & $T'$ \\
\midrule
$k$NN-MT & 0.7 & 10 & --- & --- & 0.8 & 10 & --- & --- & 0.7 & 10 & --- & --- & 0.6 & 100 & --- & ---  \\
Efficient $k$NN-MT  & 0.7 & 10 & --- & --- & 0.8 & 10 & --- & --- & 0.7 & 10 & --- & --- & 0.7 & 100 & --- & --- \\
Chunk-based $k$NN-MT & 0.7 & 10 & 0.5 & 1 & 0.8 & 10 & 0.5 & 1 & 0.7 & 10 & 0.4 & 2 & 0.8 & 100 & 0.4 & 3 \\
\midrule
\textbf{Ablations}\\
\midrule
$\textbf{c=6}$ \textbf{,} $\textbf{i=6}$\\
Maintain order & 0.8 & 10 & 0.4 & --- & 0.7 & 10 & 0.4 & --- & 0.5 & 10 & 0.4 & --- & 0.7 & 100 & 0.4 & ---\\
Single chunk   & 0.8 & 10 & 0.4 & 2 & 0.7 & 10 & 0.4 & 2 & 0.7 & 10 & 0.4 & 2 & 0.7 & 100 & 0.4 & 3   \\
All chunks     & 0.7 & 10 & 0.5 & 1 & 0.8 & 10 & 0.5 & 1 & 0.7 & 10 & 0.4 & 2 & 0.6 & 100 & 0.4 & 1  \\
Keep previous  & 0.7 & 10 & 0.5 & 1 & 0.8 & 10 & 0.5 & 1 & 0.7 & 10 & 0.4 & 1 & 0.7 & 100 & 0.4 & 3   \\
\midrule
\textbf{Keep previous} \\
$i=6$ & 0.7 & 10 & 0.5 & 1 & 0.8 & 10 & 0.5 & 1 & 0.7 & 10 & 0.4 & 1 & 0.7 & 100 & 0.4 & 3   \\
$i=8$ & 0.7 & 10 & 0.4 & 1 & 0.8 & 10 & 0.5 & 1 & 0.7 & 10 & 0.4 & 1 & 0.7 & 100 & 0.4 & 3   \\
$\exp(i_{\max}=16)$ & 0.7 & 10 & 0.5 & 1 & 0.8 & 10 & 0.5 & 1 & 0.7 & 10 & 0.4 & 2 & 0.8 & 100 & 0.4 & 3 \\
$\exp(i_{\max}=32)$ & 0.8 & 10 & 0.5 & 1 & 0.6 & 10 & 0.5 & 1 & 0.6 & 10 & 0.4 & 1 & 0.5 & 100 & 0.4 & 1   \\

\bottomrule
\end{tabular}
\caption{Values of the  hyperparameters: number of neighbors to be retrieved $k$, interpolation coefficient $\lambda$, and retrieval softmax temperature $T$.}
\label{table:hyperparameters}
\end{table*}

\begin{table*}[hptb!]
\centering \small
\setlength{\tabcolsep}{2ex}
\begin{tabular}
{ccc@{\hspace{15ex}}ccc}
\toprule
\multicolumn{3}{c}{Medical} & \multicolumn{3}{c}{Law} \\
$\eta$ & schedule & warmup & $\eta$ & schedule & warmup \\
\midrule
$1\times 10^{-5}$ & on plateau & no & $5\times 10^{-5}$ & inverse sqrt & yes  \\
\midrule
\multicolumn{3}{c}{IT} & \multicolumn{3}{c}{Koran} \\
$\eta$ & schedule & warmup & $\eta$ & schedule & warmup \\
\midrule
$5\times 10^{-5}$ & inverse sqrt & yes & $5\times 10^{-5}$ & inverse sqrt & no\\
\bottomrule
\end{tabular}
\caption{Values of the hyperparameters for the fine-tuned model.}
\label{table:hyperparameters_finetuned}
\end{table*}

\section{On-the-fly Adaptation on Law Domain}
\label{sec:on_the_fly_adaptation_law}
\begin{figure*}[hbt]
    \centering
    \includegraphics[width=7.3cm]{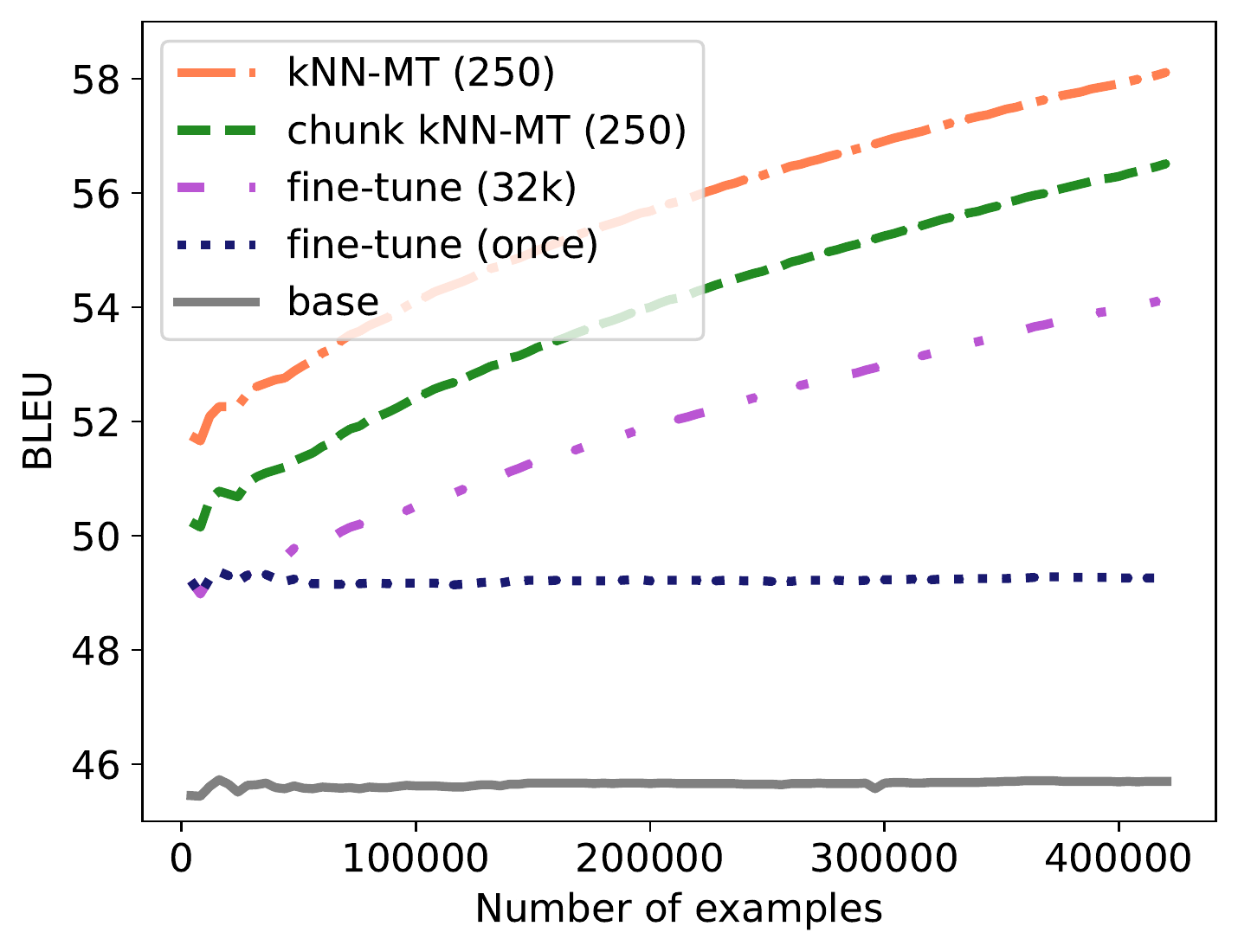} \qquad
    \includegraphics[width=7.5cm]{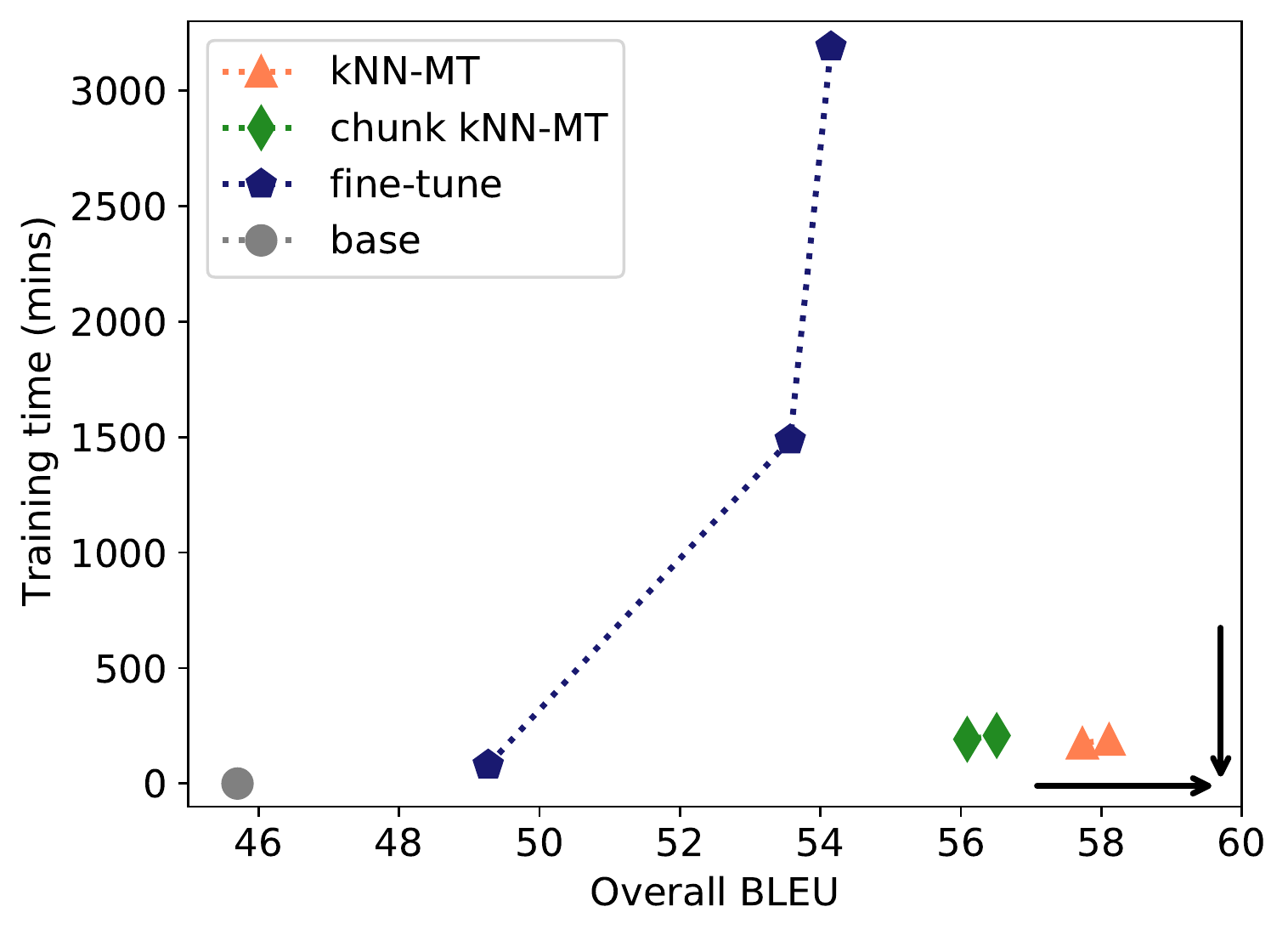}
    \includegraphics[width=7.5cm]{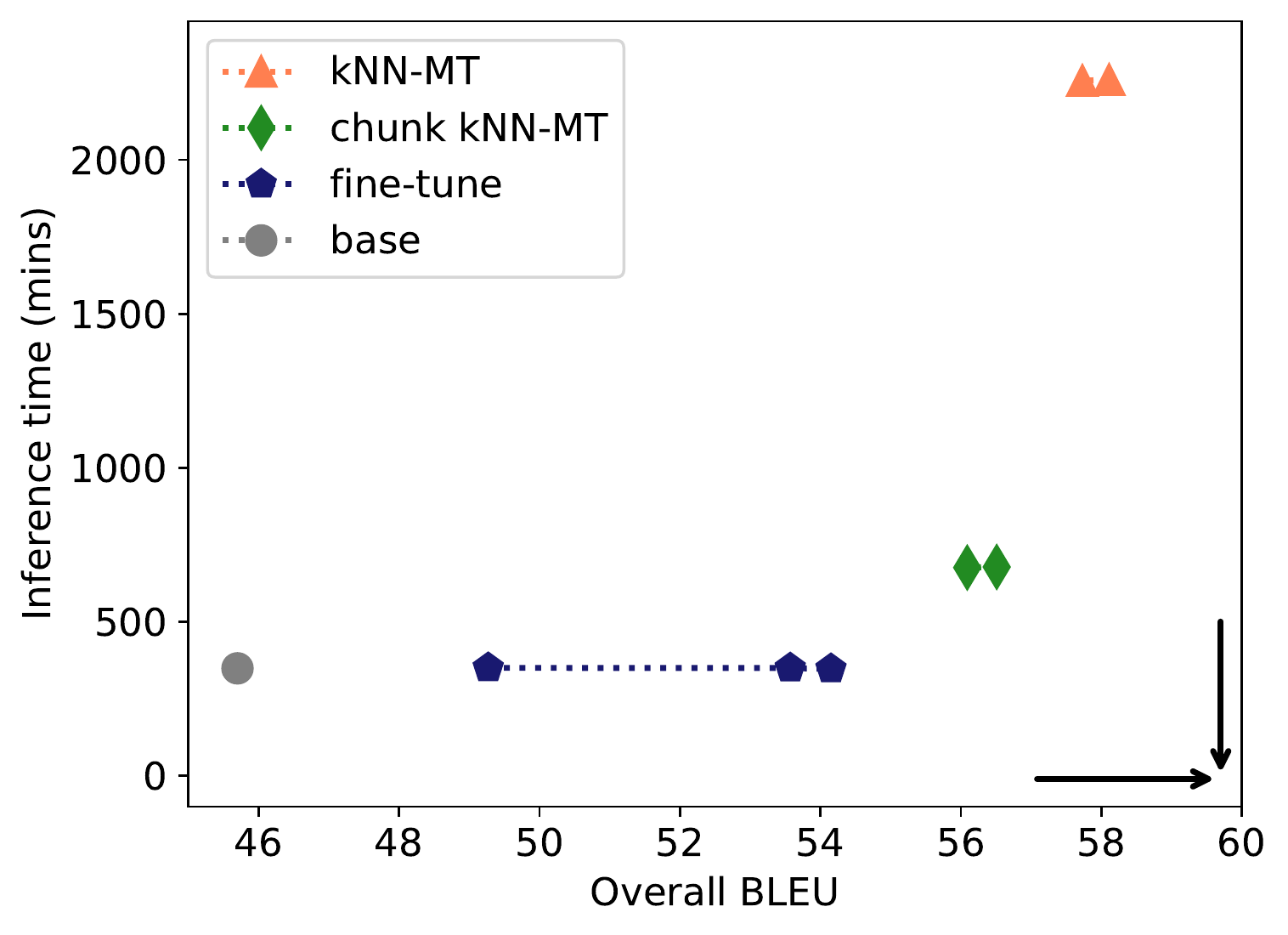} \qquad
    \includegraphics[width=7.5cm]{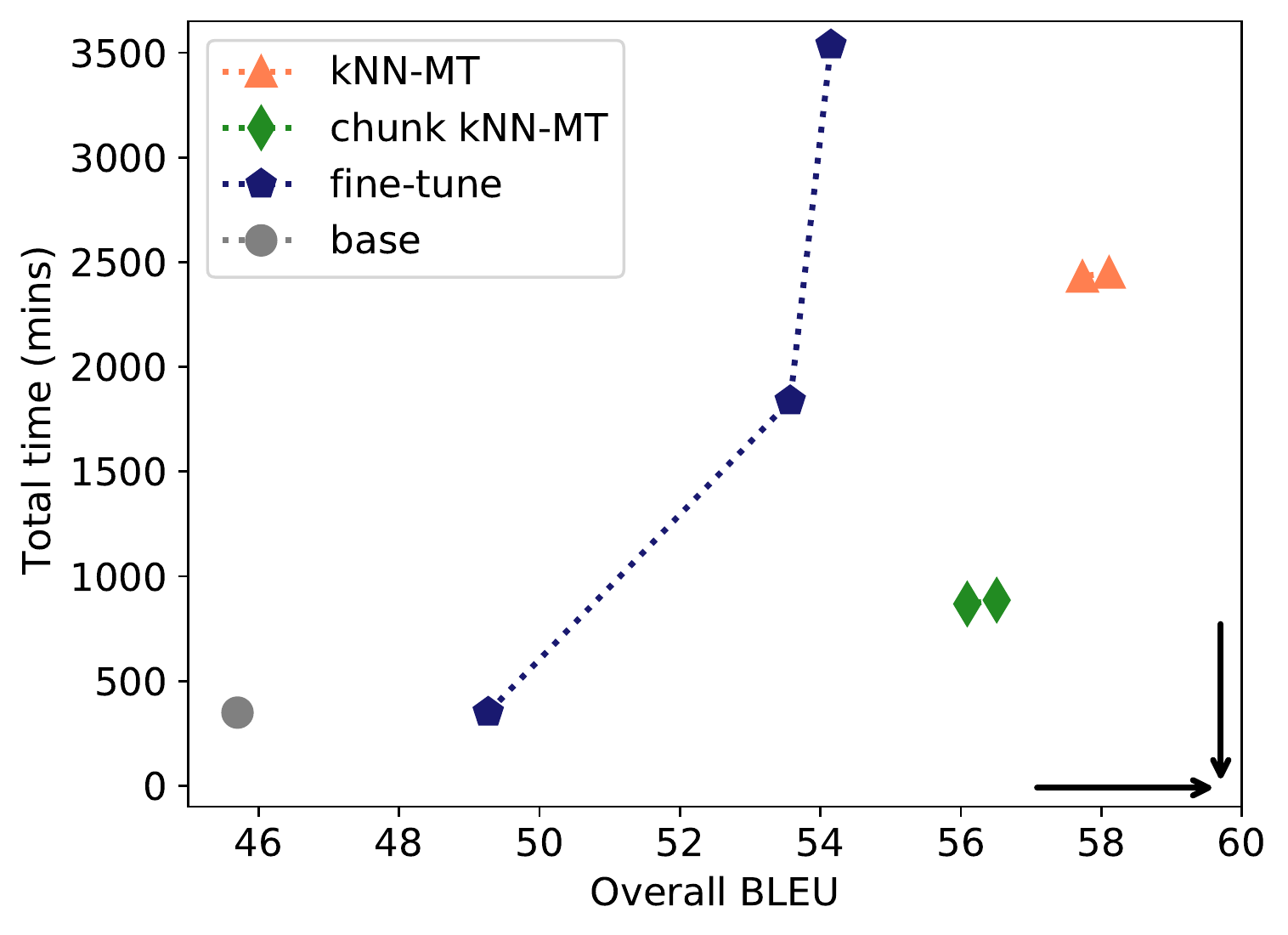}
    \caption{Analysis of the on-the-fly adaptation experiments on the law domain. Top left: BLEU scores measured every 4,000 examples. Top right: Time (in minutes) spent by each model on training / creating and updating the datastore and the corresponding BLEU score. Bottom left: Inference time (in minutes) spent by each model to translate the whole set of examples and its BLEU score. Bottom right: Total time ($\text{training} + \text{inference}$) spent by each model to translate the set of examples and its BLEU score. The lower right corner is better.}
    \label{fig:results_on_the_fly_adaptation_law}
\end{figure*}

We report the results of the on-the-fly adaptation experiment, on the law domain, on Figure \ref{fig:results_on_the_fly_adaptation_law}. 
In a similar way as in the medical domain, the top left plot shows that the $k$NN-MT and the chunk-based $k$NN-MT models lead to higher BLEU scores than the fine-tuned models. 
We can also see, on the top right plot, that the time the models take to add examples to the datastore along the generation is much shorter than the time needed to fine-tune the model. 
This comes at the cost of a higher inference time (as shown on the bottom left plot). However, we can see that the chunk-based $k$NN-MT model is able to substantially reduce the inference time gap between fully-parametric and semi-parametric models, having a shorter total time than the $k$NN-MT and the models fine-tuned every 32,000 and 64,000 steps (bottom right plot).
Concerning the fine-tuned models, fine-tuning more often leads to a slightly better BLEU score. However, this also leads to a substantially higher training time.

\section{Translation Examples}
\label{sec:examples}
\begin{figure*}[hptb!]
    \centering
    \includegraphics[width=\textwidth]{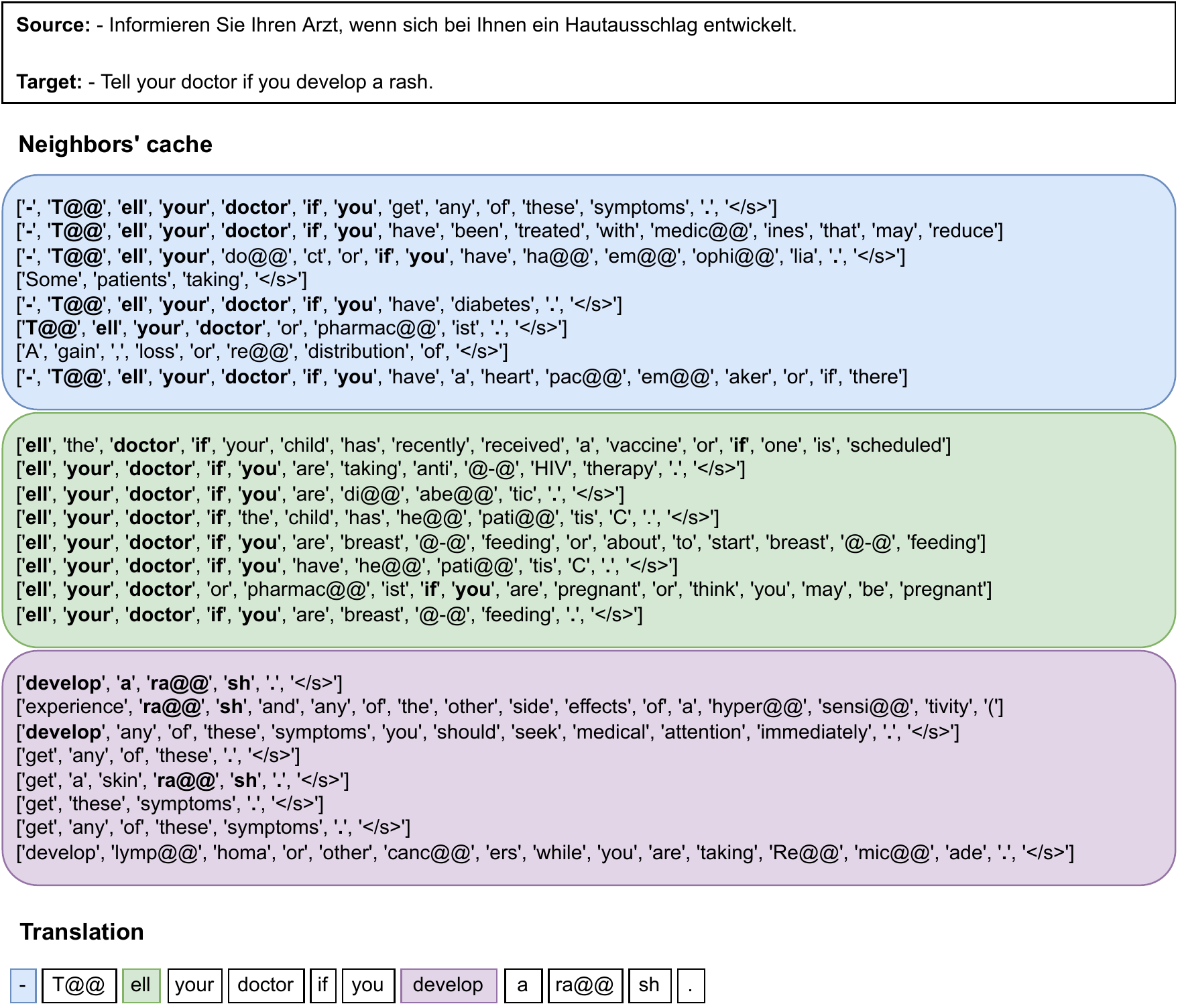}
    \caption{Example of translation from the medical domain. The tokens generated at the retrieval steps are highlighted in the same color as the retrieved chunks of tokens. These chunks of tokens are added to the neighbors' cache, after being retrieved. The retrieved tokens which are present in the translation are bolded.}
    \label{fig:example_medical_1}
\end{figure*}

\begin{figure*}[hptb!]
    \centering
    \includegraphics[width=\textwidth]{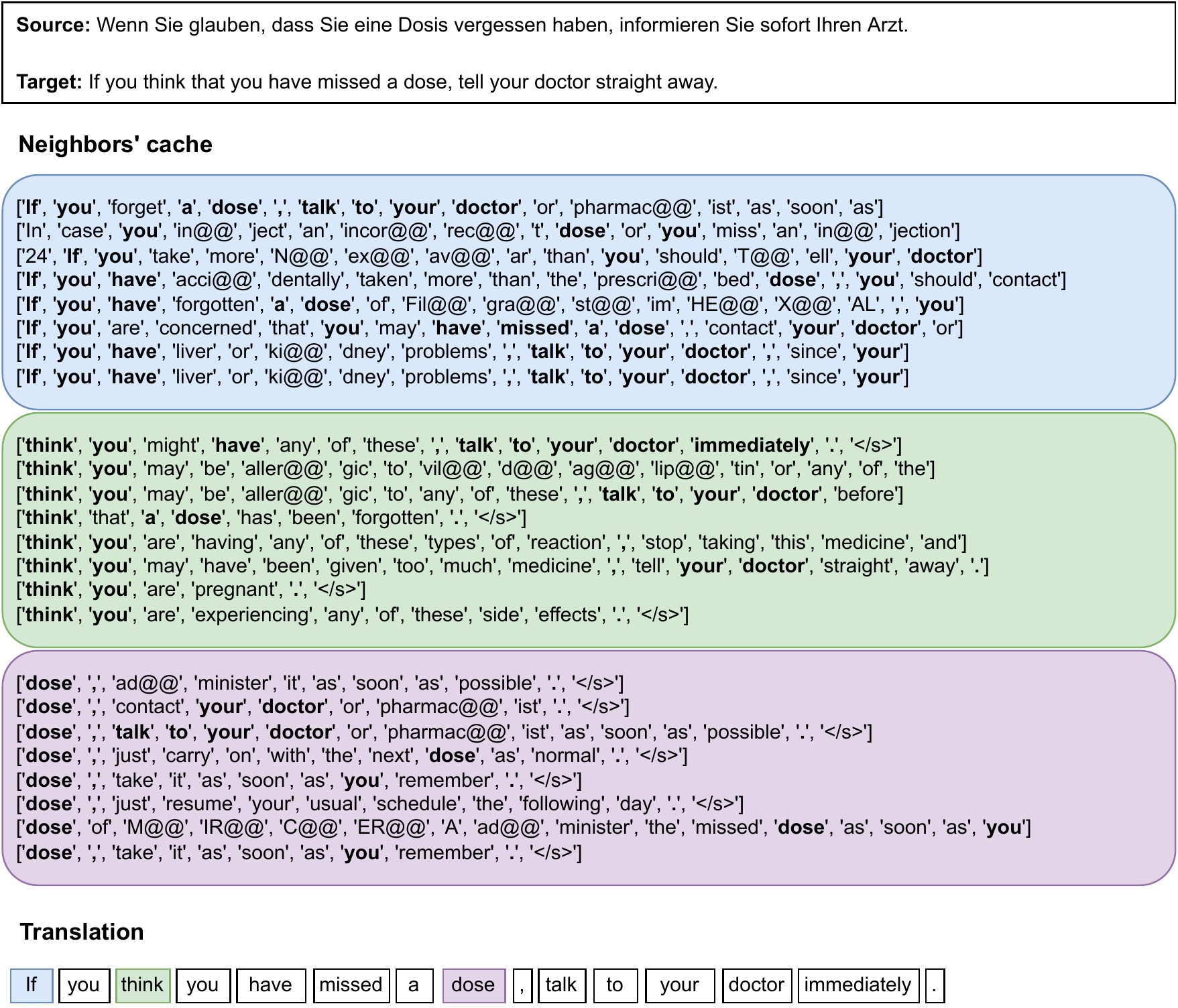}
    \caption{Example of translation from the medical domain. The tokens generated at the retrieval steps are highlighted in the same color as the retrieved chunks of tokens. These chunks of tokens are added to the neighbors' cache, after being retrieved. The retrieved tokens which are present in the translation are bolded.}
    \label{fig:example_medical_2}
\end{figure*}

\begin{figure*}[hptb!]
    \centering
    \includegraphics[width=\textwidth]{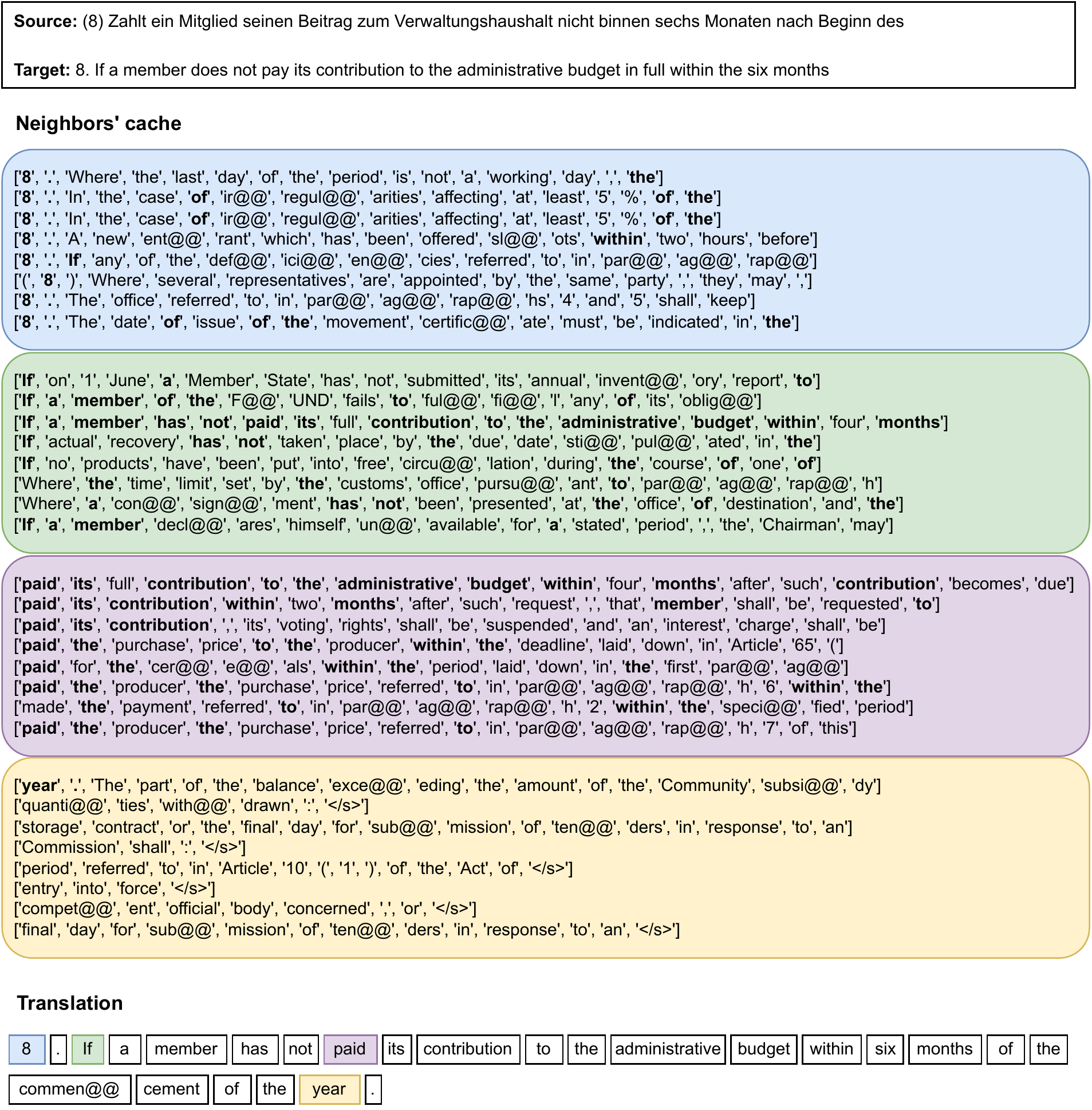}
    \caption{Example of translation from the law domain. The tokens generated at the retrieval steps are highlighted in the same color as the retrieved chunks of tokens. These chunks of tokens are added to the neighbors' cache, after being retrieved. The retrieved tokens which are present in the translation are bolded.}
    \label{fig:example_law_1}
\end{figure*}

\begin{figure*}[hptb!]
    \centering
    \includegraphics[width=\textwidth]{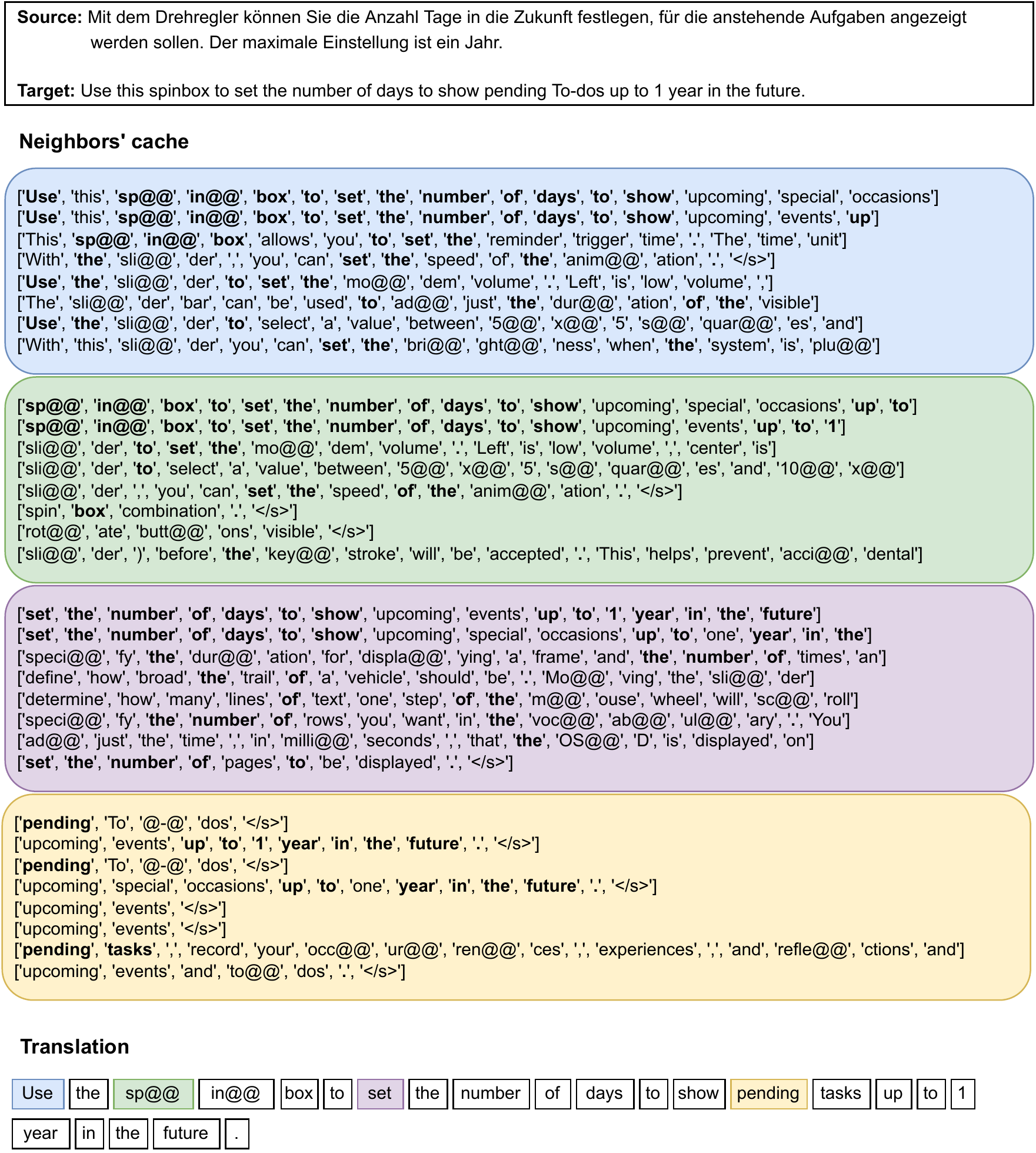}
    \caption{Example of translation from the IT domain. The tokens generated at the retrieval steps are highlighted in the same color as the retrieved chunks of tokens. These chunks of tokens are added to the neighbors' cache, after being retrieved. The retrieved tokens which are present in the translation are bolded.}
    \label{fig:example_it_1}
\end{figure*}

We report some translation examples on the medical domain in Figures \ref{fig:example_medical_1} and \ref{fig:example_medical_2}, on the law domain in Figure \ref{fig:example_law_1}, and on the IT domain in Figure \ref{fig:example_it_1}. To simplify the examples, we use a batch size of 1 and a beam size of 1.

\end{document}